\documentclass{article} % For LaTeX2e
\usepackage{iclr2022_conference,times}

\usepackage{amsmath,amsfonts,bm}

\def\eqref#1{equation~\ref{#1}}
\def\1{\bm{1}}

\DeclareMathAlphabet{\mathsfit}{\encodingdefault}{\sfdefault}{m}{sl}
\SetMathAlphabet{\mathsfit}{bold}{\encodingdefault}{\sfdefault}{bx}{n}

\newcommand{\E}{\mathbb{E}}

\usepackage{hyperref}
\usepackage{url}

\usepackage{amsmath, amssymb, amsthm, amsbsy}
\usepackage{bm}
\usepackage{graphicx}
\usepackage{xspace}
\usepackage{subcaption}
\usepackage{url}
\usepackage{enumerate}
\usepackage{sidecap}
\usepackage{standalone}
\usepackage{caption}
\usepackage{multirow}
\usepackage{dsfont}
\usepackage[subtle]{savetrees}
\usepackage{booktabs} 
\usepackage{colortbl} 
\usepackage{xfrac}
\usepackage{stackengine}
\usepackage{scalerel}

\stackMath
\def\dyhat{.067ex}
\newcommand\myhat[1]{\ThisStyle{%
              \stackon[\dyhat]{\SavedStyle#1}
                              {\SavedStyle\hat{\phantom{#1}}}}}

\newcommand{\xvec}{\mathbf{x}}
\newcommand{\zvec}{\mathbf{z}}
\newcommand{\yvec}{\mathbf{y}}
\newcommand{\nvec}{\mathbf{n}}

\usepackage{wasysym}
\newtheorem{theorem}{Theorem}
\newtheorem{definition}{Definition}
\newtheorem{lemma}{Lemma}

\newtheorem{proposition}{Proposition}

\newcommand{\stacklabel}[1]{\stackrel{\smash{\scriptscriptstyle \mathrm{#1}}}}
\newcommand{\defeq}{\stacklabel{def}=}

\title{Consistent Counterfactuals for Deep Models}

\iclrfinalcopy
\author{Emily Black\thanks{Equal Contribution} \\
Department of Computer Science\\
Carnegie Mellon University\\
Pittsburgh, PA 15213, USA \\
\texttt{emilybla@cs.cmu.edu} 
\And $\text{Zifan Wang}^*$ \\
Department of Electrical and Computer Engineering\\
Carnegie Mellon University\\
Pittsburgh, PA 15213, USA  \\
\texttt{zifanw@cmu.edu} 
\And Matt Fredrikson \\
Department of Computer Science\\
Carnegie Mellon University\\
Pittsburgh, PA 15213, USA  \\
\texttt{mfredrik@cs.cmu.edu} 
\And Anupam Datta \\
Department of Electrical and Computer Engineering\\
Carnegie Mellon University\\
Mountain View, CA 94043, USA  \\
\texttt{danupam@cmu.edu} 
}

\begin{document}

\maketitle

\begin{abstract}
   Counterfactual examples are one of the most commonly-cited methods for explaining the predictions of machine learning models in key areas such as finance and medical diagnosis.  Counterfactuals are often discussed under the assumption that the model on which they will be used is static, but in deployment models may be periodically retrained or fine-tuned. This paper studies the consistency of model prediction on counterfactual examples in deep networks under small changes to initial training conditions, such as weight initialization and leave-one-out variations in data, as often occurs during model deployment. We demonstrate experimentally that counterfactual examples for deep models are often inconsistent across such small changes, and that increasing the cost of the counterfactual, a stability-enhancing mitigation suggested by prior work in the context of simpler models, is not a reliable heuristic in deep networks. Rather, our analysis shows that a model's Lipschitz continuity around the counterfactual, along with confidence of its prediction, is key to its consistency across related models. To this end, we propose Stable Neighbor Search as a way to generate more consistent counterfactual explanations, and illustrate the effectiveness of this approach on several benchmark datasets.
    
\end{abstract}
% !TEX ROOT=./paper.tex

\section{Introduction}\label{sec:introduction}
% Deep Neural Networks (DNNs) are increasingly being integrated into important decision-making processes, from medical diagnosis~\citep{} to credit risk analysis~\citep{}. In several of these applications, it is necessary to generate explanations of a deployed model's decision: either for the model's user to understand the decision well enough to trust it, such as in a medical context; or it even may be legally required,  such as in a financial loan application context, in order for the decision recipient learn how to change their behavior in order to get a different outcome from the model in the future.
Deep Networks are increasingly being integrated into decision-making processes which require explanations during model deployment, from medical diagnosis to credit risk analysis~\citep{Bakator_2018,litjens2017medical,liu2014early,sun2016computer,de2018clinically, babaev2019rnn,Addo_2018,balasubramanian2018insurance,wang2018leveraging}.
%hiring~\citep{}--- and are on the horizon for myriad more applications, such as credit risk analysis~\citep{}. 
%Many of these applications require that a deployed model generate explanations of its decisions---for example, it is legally required that rejected loan applicants will receive an explanation of their decision if requested. %Even if it is not legally mandated, it is increasingly considered necessary for models integrated in important decision systems, such as those in medicine, to return explanations as required as a part of their deployment in order to increase user trust and understanding~\citep{PWC}. 
\emph{Counterfactual examples}~\citep{wachter2018counterfactual,van2019interpretable,mahajan2019preserving,verma2020counterfactual,laugel2017inverse,keane2020good,ustun2019actionable,sharma2019certifai,poyiadzi2020face,karimi2020algorithmic,pawelczyk2020learning} %are a common and well-studied explanation method that 
are often put forth as a simple and intuitive method of explaining decisions in such high-stakes contexts~\citep{mc2018interpretable,yang2020generating}. A counterfactual example for an input $x$ is a related point $x'$ that produces a desired outcome $y'$ from a model. % (e.g., the nearest potential accepted loan application to a rejected loan application). 
Intuitively, these explanations are intended to answer the question, ``Why did point $x$ not receive outcome $y'$?'' either to give instructions for \emph{recourse}, i.e. how an individual can change their behavior to get a different model outcome, or as a check to ensure a model's decision is well-justified~\citep{ustun2019actionable}. 

Counterfactual examples are often viewed under the assumption that the decision system on which they will be used is static: that is, the model that \emph{creates} the explanation will be the \emph{same} model to which, e.g. a loan applicant soliciting recourse re-applies~\citep{barocas2016big}. %, or that the doctor requiring an justification for a model's decision decides to trust. 
However, during real model deployments in high-stakes situations, models are not constant through time: there are often retrainings due to small dataset updates, or fine-tunings to ensure consistent good behavior~\citep{merchant_2020,pwc}. Thus, in order for counterfactuals to be usable in practice, they must return the same desired outcome not only for the model that generates them, but for similar models created during deployment. 
This paper investigates the consistency of model predictions on counterfactual examples between deep models with seemingly inconsequential differences, i.e. random seed and one-point changes in the training set. We demonstrate that some of the most common methods generating counterfactuals in deep models either are highly inconsistent between models or very costly in terms of distance from the original input. Recent work that has investigated this problem in simpler models~\citep{pmlr-v124-pawelczyk20a} has pointed to increasing counterfactual cost, i.e. the distance between an input point and its counterfactual, as a method of increasing consistency. We show that while higher than \emph{minimal} cost is necessary to achieve a stable counterfactual, cost alone is not a reliable signal to guide the search for stable counterfactuals in deep models (Section ~\ref{sec:invalidation}). 

Instead, we show that a model's Lipschitz continuity and confidence around the counterfactual is a more reliable indicator of the counterfactual's stability. 
Intuitively, this is due to the fact that these factors bound the extent of a models local decision boundaries will change across fine-tunings, which we prove in Section ~\ref{sec:method}.
%Intuitively, this is due to the fact that points in neighborhoods with small Lipschitz constants are more consistent across model fine-tunings, which we prove in Section ~\ref{sec:method}. 
Following this result, we introduce \emph{Stable Neighbor Search} (SNS), which finds counterfactuals by searching for high-confidence points with small Lipschitz constants in the generating model (Section~\ref{sec:method}). Finally, we empirically demonstrate that SNS generates consistent counterfactuals while maintaining a low cost relative to other methods over several tabular datasets, e.g. Seizure and German Credit from UCI database~\citep{uci}, in Section ~\ref{sec:evaluation}.
% Thus, as we empirically demonstrate (Section~\ref{sec:invalidation}), increasing a counterfactual's cost is does not guarantee higher consistency between nearby models.
% As our empirically results show
% As we later demonstrate (Section ~\ref{sec:evaluation}), our method, Robust Neighbor Search provides a principled way to generate consistent counterfactuals at a lower cost (less than half) than solutions recommended by previous work.
%Thus, as we empirically demonstrate~\ref{sec:evaluation}, naively increasing a counterfactual's cost as a method of increasing consistency between nearby models can either be ineffective, or lead to counterfactuals with extremely high costs.  
%, and finally, empirically demonstrate its ability to find stable counterfactuals at a relatively low cost compared to other methods (Section~\ref{sec:evaluation}). 
% SNS hinges on the theoretical observation that points with a low Lipschitz constant in the model are more difficult for models to change their prediction on during training (Section ~\ref{sed:method}). %This is due to the fact that, from the perspective of the loss, these points incur a high cost to move between classes. 
% This suggests that they the outcome of such points will be more consistent between models with small differences, as they optimize the same loss. Following this intuition, we prove that RNS finds counterfactuals by searching for points with a low local Lipschitz constant in the generating model (Section~\ref{sec:method}).

In summary, our main contributions are: 1) we demonstrate that common counterfactual explanations can have low consistency across nearby \emph{deep} models, and that cost is an insufficient signal to find consistent counterfactuals (Theorem.~\ref{theorem:cost-in-deep-models}); 2) to navigate this cost-consistency tradeoff, we prove that counterfactual examples in a neighborhood where the network has a small local Lipschitz constant are more consistent against small changes in the training environment (Theorem.~\ref{theorem:loss-and-lipschitz}); 3) leveraging this result, we propose SNS as a way to generate consistent counterfactual explanations (Def.~\ref{def:SNS}); 4) we empirically demonstrate the effectiveness of SNS in generating consistent and low-cost counterfactual explanations (Table~\ref{table_IV}).
More broadly, this paper further develops a connection between the geometry of deep models and the consistency of counterfactual examples.
When considered alongside related findings that focus on attribution methods, our work adds to the perspective that \emph{good explanations require good models to begin with}~\citep{croce2019provable,wang2020smoothed,dombrowski2019explanations,simonyan2013deep,sundararajan2017axiomatic}.

\section{Background}\label{sec:related-work}
\paragraph{Notation.} We begin with notation, preliminaries, and definitions. %an introduction to the counterfactual examples to be discussed in this paper.
Let $F(\xvec;\theta) = \arg\max_i f_i(\xvec;\theta)$ be a deep network where $f_i$ denotes the logit output for the $i$-th class and $\theta$ is the vector of trainable parameters. If $F(\xvec;\theta) \in \{0, 1\}$, there is only one logit output so we write $f$. Throughout the paper we assume $F$ is piece-wise linear such that all the activation functions are ReLUs. We use $||\xvec||_p$ to denote the $\ell_p$ norm of a vector $\xvec$ and $B_p(\xvec, \epsilon) \defeq \{\xvec' | ||\xvec'-\xvec||_p \leq \epsilon, \xvec' \in \mathbb{R}^d \}$ to denote a norm-bounded ball around $\xvec$.

\paragraph{Counterfactual Examples.} We introduce some general notation to unify the definition of a counterfactual example across various approaches with differing desiderata. In the most general sense, a counterfactual example for an input $\xvec$ is an example $\xvec_c$ that receives the different, often targeted, prediction while minimizing a user-defined \emph{quantity of interest} (QoI) (see Def.~\ref{def:counterfactual-example}): for example, a counterfactual explanation for a rejected loan application is a related hypothetical application that was accepted. We refer to the point $\xvec$ requiring a counterfactual example the \emph{origin point} or \emph{input} interchangeably.

\begin{definition}[Counterfactual Example]\label{def:counterfactual-example}
    Given a model $F(\xvec)$, an input $\xvec$, a desired outcome class $c \neq F(\xvec;\theta)$ ,
    and a user-defined quantity of interest $q$,
     a counterfactual example $\xvec_c$ for $\xvec$ is defined as $\xvec_c \defeq \arg\min_{F(\xvec';\theta) =c} q(\xvec',\xvec)$ where the \emph{cost} of $\xvec_c$ is defined as $||\xvec-\xvec_c||_p$.
\end{definition}
\label{def:counterfactual}

%Counterfactual explanations are a method to either understand a model's decision on a given point, or provide clues for how an individual subject to a model's decision may change their behavior (and thus their input to the model) to get a different result, i.e. to provide recourse. 

The majority of counterfactual generation algorithms minimize of $q_{\text{low}}(\xvec, \xvec')\defeq ||\xvec-\xvec'||_p$, potentially along with some constraints, to encourage low-cost counterfactuals~\citep{wachter2018counterfactual}. Some common variations include ensuring that counterfactuals are attainable, i.e. not changing features that cannot be changed (e.g. sex, age) due to domain constraints~\citep{ustun2019actionable,lash2017generalized}, ensuring sparsity, so that fewer features are changed~\citep{dandl2020multi,guidotti2018local}, or incorporating user preferences into what features can be changed ~\citep{mahajan2019preserving}. Alternatively, a somewhat distinct line of work~\citep{pawelczyk2020learning,van2019interpretable,joshi2019towards} also adds constraint to ensure that counterfactuals come from the data manifold. %to enforce Def.~\ref{def:counterfactual-example} returns an counterfactual $\xvec_c$ on a known data manifold $\mathcal{D}$. 
Other works still integrate causal validity into counterfactual search~\citep{karimi2020algorithmic}, or generate multiple counterfactuals at once~\citep{mothilal2020explaining}. 

%In the rest of the paper, we follow Pawelczyk et al in restricting our analysis to the first two approaches.%focusing on the first two approaches, and comparing to samples from each. 
We focus our analysis on the first two approaches, which we denote \emph{minimum-cost} and \emph{data-support} counterfactuals. 
We make this choice as the causal and distributional assumptions used in other counterfactual generation methods referenced are specific to a given application domain, whereas our focus is on the general properties of counterfactuals across domains. 
%Given that our focus is on the general properties of counterfactuals across domains, whereas approaches that require outside knowledge, such as a causal graph, user preferences over which features to change, or models of what features may or may not be mutable, are always specific to a given application domain. 
Specifically, we evaluate our results on minimum-cost counterfactuals introduced by ~\citet{wachter2018counterfactual}, and data-support counterfactuals from ~\citet{pawelczyk2020learning}, and ~\citet{van2019interpretable}. 
We give the full descriptions of these approaches in Sec.~\ref{sec:evaluation}.

\paragraph{Counterfactual Consistency.} Given two models $F(\xvec;\theta_1)$ and $F(\xvec;\theta_2)$, a counterfactual example $\xvec_c$ for $F(\xvec;\theta_1)$ is consistent with respect to $F(\xvec;\theta_2)$ means $F(\xvec_c;\theta_1) = F(\xvec_c;\theta_2)$. Following~\citet{pmlr-v124-pawelczyk20a}, we define the \emph{Invalidation Rate} for counterfactuals in Def.~\ref{def:invalidation-rate}.

\begin{definition}[Invalidation Rate]\label{def:invalidation-rate}
    Suppose $\xvec_c$ is a counterfactual example for $\xvec$ found in a model $F(\xvec;\theta)$, we define the invalidation rate \text{\emph{IV}($\xvec_c, \Theta$)} of $\xvec_c$ with respect to a distribution $\Theta$ of trainable parameters as $\emph{IV}(\xvec_c, \Theta) \defeq \mathbb{E}_{\theta' \sim \Theta} \mathbb{I}[F(\xvec_c;\theta') \neq F(\xvec_c;\theta)]$.
\end{definition}
Throughout this paper, we will call the model $F(\xvec;\theta)$ that creates the counterfactual the \emph{generating} or \emph{base} model. Recent work has investigated the consistency of counterfactual examples across similar linear and random forest models~\citep{pmlr-v124-pawelczyk20a}. 
We study the invalidation rate with respect to the distribution $\Theta$ introduced by arbitrary differences in the training environment, such as random initialization and one-point difference in the training dataset. %We further restrict all models $F(\xvec;\theta'), \theta' \sim \Theta$ have similar classification performance as the generating model $F(\xvec;\theta)$. That is, the test accuracy of $F(\xvec;\theta')$ is within a certain accuracy bound of $F(\xvec;\theta)$. 
We also assume $F(\xvec;\theta')$ uses the same set of hyper-parameters as chosen for $F(\xvec;\theta)$, e.g. the number of epochs, the optimizer, the learning rate scheduling, loss functions, etc.

%!TEX root=./paper.tex

\section{Counterfactual Invalidation in Deep Models}\label{sec:invalidation}

As we demonstrate in more detail in Section ~\ref{sec:evaluation}, counterfactual invalidation is a problem in deep networks on real data: counterfactuals produce inconsistent outcomes in duplicitous deep models up to 94\% of the time. 
%emily: the average is 40% of the time. Should we put that in instead?

Previous work investigating the problem of counterfactual invalidation~\citep{pmlr-v124-pawelczyk20a,rawal2021i}, has pointed to increasing counterfactual cost as a potential mitigation strategy.
%Previous work has pointed out the relationship between counterfactual cost and invalidation rate, 
In particular, they prove that higher cost counterfactuals will lead to lower invalidation rates in linear models in expectation~\citep{rawal2021i}, and demonstrate their relationship in a broader class of well-calibrated models~\citep{pmlr-v124-pawelczyk20a}. While this insight provides interesting challenge to the perspective that low cost counterfactuals should be preferred, we show that cost alone is insufficient to determine which counterfactual has a greater chance of being consistent at generation time in deep models.

The intuition that a larger distance between input and counterfactual will lead to lower invalidation rests on the assumption that the distance between a point $x$ and a counterfactual $x_c$ is indicative of the distance from $x_c$ to the decision boundary, with a higher distance making $x_c$'s prediction more stable under perturbations to that boundary. This holds well in a linear model, where there is only one boundary~\citep{rawal2021i}. 

However, in the complex decision boundaries of deep networks, going farther away from a point across the \emph{nearest} boundary may lead to being closer to a \emph{different} boundary. We prove that this holds even for a one-hidden-layer network by Theorem~\ref{theorem:cost-in-deep-models}. This observation shows that a counterfactual example that is farther from its origin point may be equally susceptible to invalidation as one closer to it. In fact, we show that the \emph{only} models where $\ell_p$ cost is universally a good heuristic for distance from a decision boundary, and therefore by the reasoning above, consistency, are linear models (Lemma~\ref{lemma:cost-in-deep-models}). 

\begin{theorem}\label{theorem:cost-in-deep-models}
Suppose that $H_1, H_2$ are decision boundaries in a piecewise-linear network $F(\xvec) = sign\{w_1^\top ReLU(W_0\xvec)\}$, and let $\xvec$ be an arbitrary point in its domain. 
If the projections of $\xvec$ onto the corresponding halfspace constraints of $H_1, H_2$ are on $H_1$ and $H_2$, then there exists a point $\xvec'$ such that:
\begin{align*}
\mathit{1)}\ d(\xvec', H_2) = 0 & &
\mathit{2)}\ d(\xvec', H_2) < d(\xvec, H_2) & &
\mathit{3)}\ d(\xvec, H_1) \leq d(\xvec', H_1)
\end{align*}
where $d(\xvec, H_*)$ denotes the distance between $\xvec$ and the nearest point on a boundary $H_*$.
\end{theorem}
    
\begin{lemma}\label{lemma:cost-in-deep-models}
Let $H_1, H_2, F$ and $\xvec$ be defined as in Theorem~\ref{theorem:cost-in-deep-models}. 
If the projections of $\xvec$ onto the corresponding halfspace constraints of $H_1, H_2$ are on $H_1$ and $H_2$, but there \emph{does not} exist a point $\xvec'$ satisfying \emph{(2)} and \emph{(3)} from Theorem~\ref{theorem:cost-in-deep-models}, then $H_1$ = $H_2$.
\end{lemma}

\begin{figure}[t]
    \centering
    \begin{subfigure}{0.29\textwidth}

        \includegraphics[width=\textwidth]{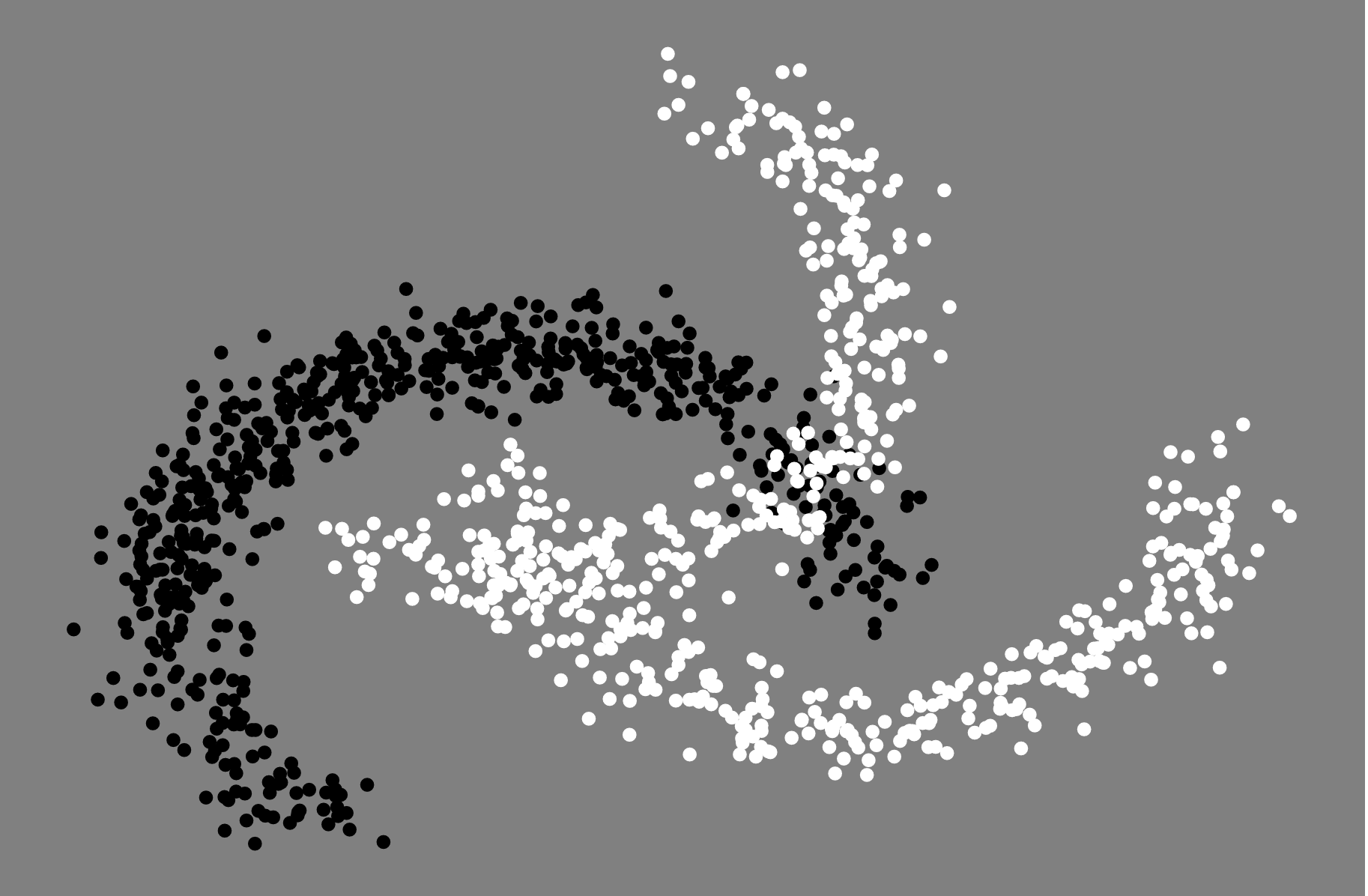}
        \caption{}
        \label{fig:toy-data}
    \end{subfigure}
    \begin{subfigure}{0.3\textwidth}

        \includegraphics[width=\textwidth]{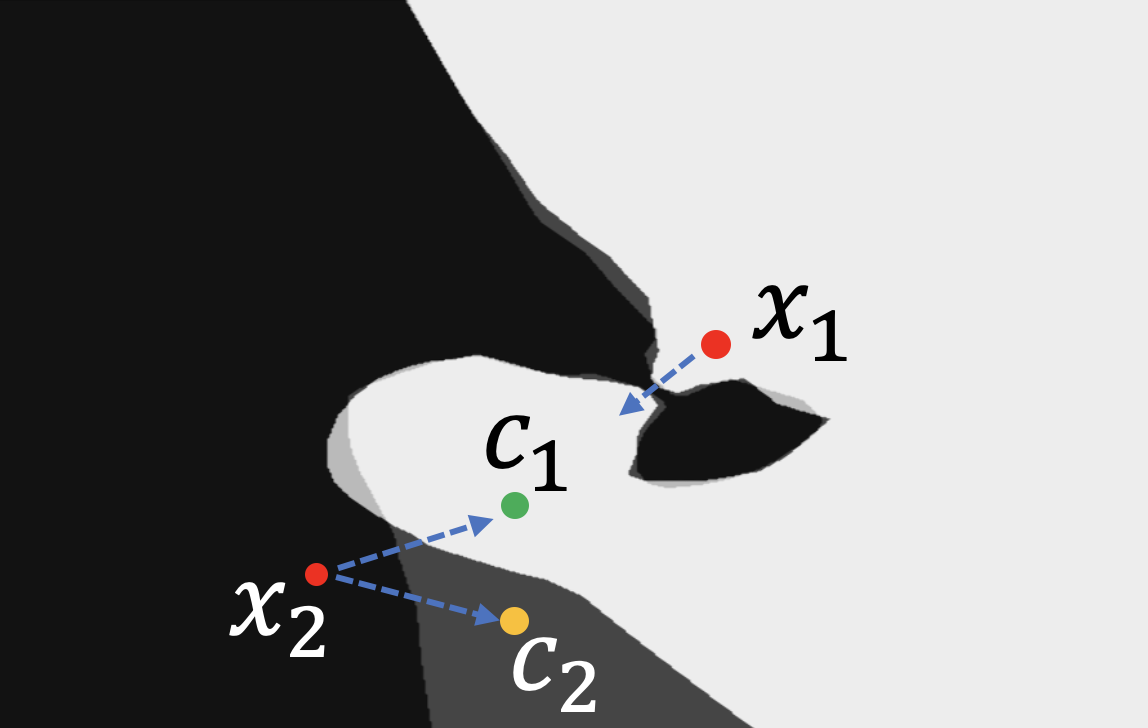}
        \caption{}
        \label{fig:toy-non-linear}
    \end{subfigure}
    \begin{subfigure}{0.3\textwidth}
        \includegraphics[width=\textwidth]{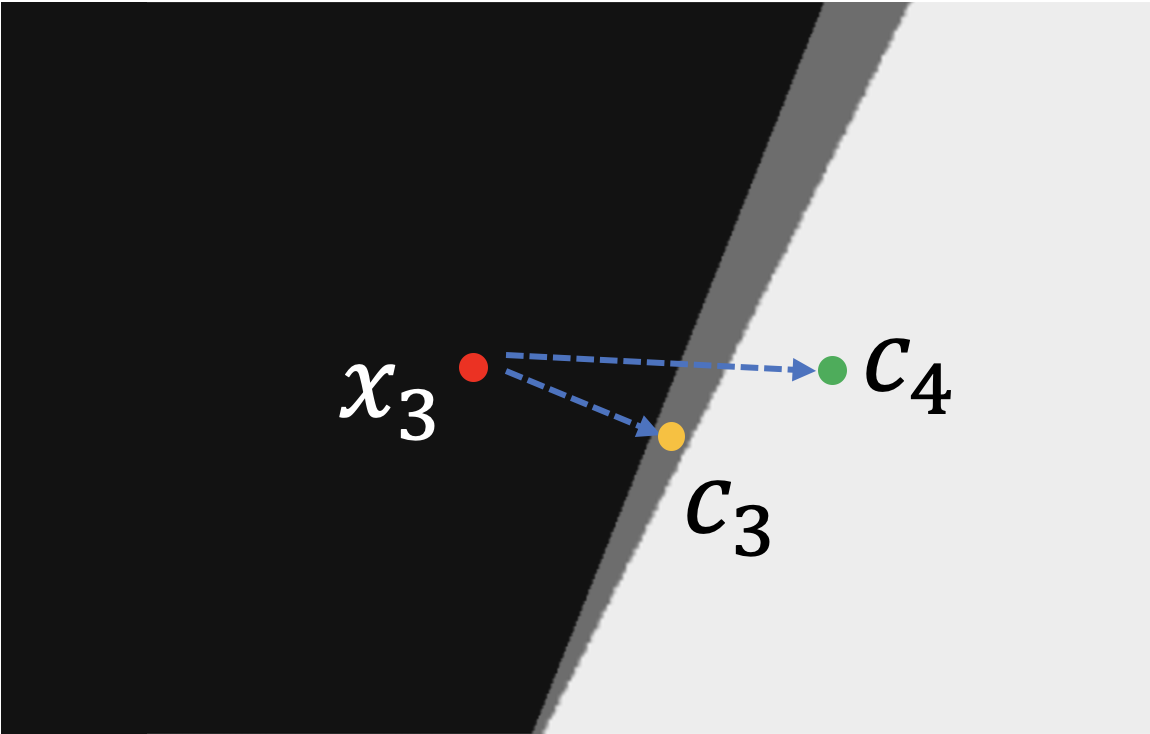}
        \caption{}
        \label{fig:toy-linear}
     \end{subfigure}
    \caption{Illustration of the boundary change in a deep model (b) and a linear model (c) for a 2D dataset (a) when changing the seed for random initialization during the training. Shaded regions correspond to the area when two deep models in (b) (or two linear models in (c)) make different predictions.}
    \label{fig:cost-intuition}
 \end{figure}

% \begin{proof}
% We prove this lemma by demonstration. Consider the deep model in Figure ~\ref{fig:lemma_pic}. See that for point $x_1$, the higher-cost counterfactual generated for that point in the is actually closer to the decision boundary than the lower-cost counterfactual.
% \end{proof}

%While Theorem~\ref{theorem:cost-in-deep-models} and Lemma~\ref{lemma:loss-and-lipschitz} are constrained in one-layer models, the insights generalize to more complicated models. 
Figure~\ref{fig:cost-intuition} illustrates the geometric intuition behind these results. 
The shaded regions of ~\ref{fig:toy-non-linear} correspond to two decision surfaces trained from different random seeds on the data in (a).
The lighter gray region denotes where the models disagree, whereas the black and white regions denote agreement.
Observe that counterfactuals equally far from a decision boundary may have different invalidation behavior, as demonstrated by the counterfactuals $c_1$ and $c_2$ for the point $x_2$.
Also note that as shown with $x_1$, being far away from one boundary may lead one to cross another one in deep models. 
However, for two linear models shown in Fig.~\ref{fig:toy-linear}, being far away from the boundary is indeed a good indicator or being consistent.

The discussion so far has demonstrated that there is not a strong theoretical relationship between cost and invalidation in deep models. 
In Section ~\ref{sec:evaluation}, we test this claim on real data, and show that higher-cost counterfactuals can have \emph{higher} invalidation rates than their lower-cost relatives (c.f. Table~\ref{table_IV}). 
Further, we show that the coefficient of determination ($R^2$) between cost and invalidation rate is very
%exceedingly
 small (with all but one around 0.05). 
Thus, while cost and invalidation are certainly related---for example, it may be necessary for a stable counterfactual to be more costly than the \emph{minimum} point across the boundary---cost alone is not enough to determine which one will be the most consistent in deep models. 
% In the next section, we show the local Lipschitz continuity of a point serves as a reliable signal to find stable counterfactuals in deep models.

\section{Towards Consistent Counterfactuals}
\label{sec:method}
In this section, we demonstrate that the Lipschitz continuity (Def.~\ref{def:lipschitz-continuity}) of a neighborhood around a counterfactual can be leveraged to characterize the consistency of counterfactual explanations under changes to the network's parameters (Section~\ref{sec:lipschitz}). Our main supporting result is given in Theorem~\ref{theorem:loss-and-lipschitz}, which shows that a model's Lipschitz constant in a neighborhood around a $x_c$ together with the confidence of its prediction on $x_c$ serve as a proxy for the difficulty of invalidating $x_c$. We further discuss insights from these analytical results and introduce an effective approach, \emph{Stable Neighbor Search}, to improve the consistency of counterfactual explanations (Section~\ref{sec:sns}).
Unless otherwise noted, this section assumes all norms are $\ell_2$. %We begin this section by defining Lipschitz Continuity.

%To reach these results, we first describe our method of modeling changes to a model's decision boundary with \emph{distributional influence}.

% \begin{definition}[Local Lipschitz Continuity]\label{def:local-lipschitz-continuity}
%     A function $h:\mathbb{R}^d \rightarrow \mathbb{R}$ is \emph{$(K, \epsilon)$-locally Lipschitz continuous at $\xvec$} iff $\forall \xvec' \in B(\xvec, \epsilon) . |h(\xvec') - h(\xvec)| \leq K ||\xvec'-\xvec||$. 
%     We refer to $K$ as the \emph{local Lipschitz constant} of $h$ at $\xvec$.
% \end{definition}

\begin{definition}[Lipschitz Continuity]\label{def:lipschitz-continuity}
    A continuous and differentiable function $h:S \rightarrow \mathbb{R}^m$ is \emph{$K$-Lipschitz continuous} iff $\forall \xvec' \in S, ||h(\xvec') - h(\xvec)|| \leq K ||\xvec'-\xvec||$. We write $h$ is $K$-Lipschitz in $S$.
\end{definition}

% \subsection{Lipschitz Continuity and Consistency}

% Characterizing the precise effect that changes such as random initialization have on the outcome of training is challenging. We approach this by modeling the differences that arise from small changes
% as a \emph{fine-tuning} of the original model, where the top layer of the model is re-trained and the parameters of rest of the layers are frozen. Within this framework, we proceed to show that given a counterfactual $x_c$ for a point $x$, the Lipschitz constant of a neighborhood around $x_c$ along with the confidence of its prediction serves as a proxy for the difficulty of invalidating $x_c$ through fine-tuning. Our main supporting result is given in Theorem~\ref{theorem:loss-and-lipschitz}, which shows that if a model's Lipschitz constant in a neighborhood around a $x_c$ as well as the confidence of its prediction on $x_c$ serve as a proxy for the difficulty of invalidating $x_c$. 

%not sure if this is good
%the change to a model's decision boundaries in an neighborhood is bounded by the Lipschitz constant in that neighborhood and the confidence of the models prediction of the counterfactual.

\subsection{ReLU Decision Boundaries and Distributional Influence}

We analyze the differences between models with changes such as random initialization by studying the differences that arise in their decision boundaries. In order to capture information about the decision boundaries in analytical form, we introduce \emph{distributional influence}: a method of using a model's gradients to gather information its local decision boundaries.
We begin motivating this choice by reviewing key aspects of the geometry of ReLU networks.
%We analyze the differences between models with changes such as random initialization by studying the differences that arise in their decision boundaries. To capture decision boundaries in analytical form, we introduce \emph{distributional influence}: method of capturing information about a model's decision boundaries using its gradients.% of using a model's gradients to gather information about the local decision boundaries in a given neighborhood.

%We characterize the effects of changes such as random initialization have on models by modeling the differences that arise from the change of the decision boundaries. In order to capture the decision boundaries in analytical form, we introduce distributional influence.
%We approach the change of model during retraining by modeling the differences that arise from the change of the decision boundaries. In order to capture the decision boundaries in analytical form, we introduce distributional influence.

% A linear model $f(\xvec) = \text{sign}[\mathbf{w}^\top\xvec+b]$ is characterized by a single a decision boundary is a linear constraint that corresponds to $f(\xvec) \geq 0$. 
ReLU networks have piecewise linear boundaries that are defined by the status of each ReLU neuron in the model~\citep{Jordan2019ProvableCF, DBLP:conf/nips/HaninR19}. 
To see this, let $u_i(\xvec)$ denote the pre-activation value of the neuron $u_i$ in the network $f$ at $\xvec$. 
% A neuron is \emph{activated} if $u_i(\xvec) \geq 0$. Then the activation status of the neuron $u_i$ is defined by $\text{sign}[u_i(\xvec)]$. 
We can associate a half-space $A_i$ in the input space with the linear activation constraint $u_i(\xvec) \geq 0$ corresponding to the \emph{activation status} of neuron $u_i$, and an $\emph{activation pattern}$ for a network at $\xvec$, $p(\xvec)$, as the activation status of every neuron in the network.
An \emph{activation region} for a given activation pattern $p$, denoted $\mathcal{R}(p)$, is then a subspace of the network's input that yields the activations in $p$; geometrically, this is a polytope given by the convex intersection of all the half-spaces described by $p$, with facets corresponding to each neuron's activation constraint.

\begin{figure}[t]
    \centering
    \begin{subfigure}[b]{0.45\textwidth}
        \centering
        \includegraphics[width=0.6\textwidth]{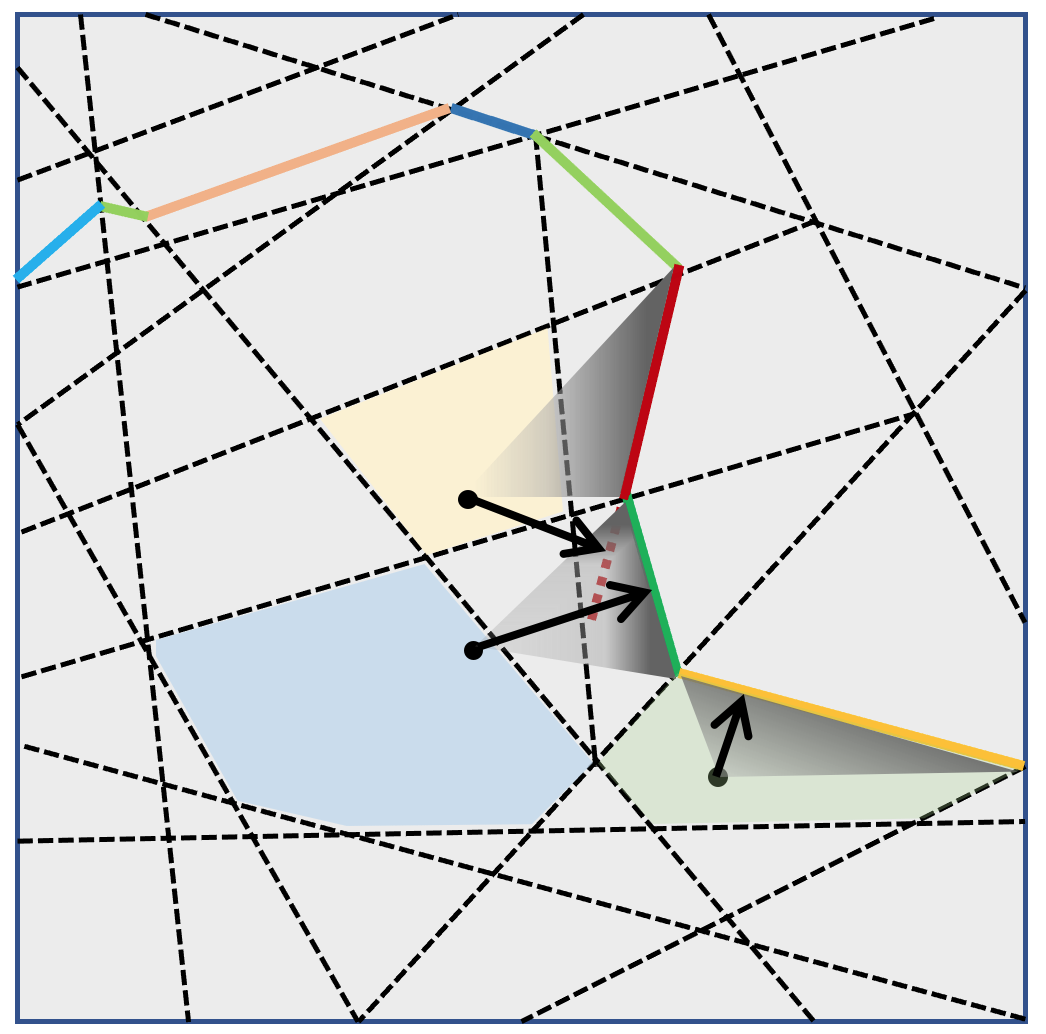}
        \caption{}
        \label{fig:illustration-circle}
    \end{subfigure}
    \hfill
    \begin{subfigure}[b]{0.45\textwidth}
        \centering
        \includegraphics[width=0.7\textwidth]{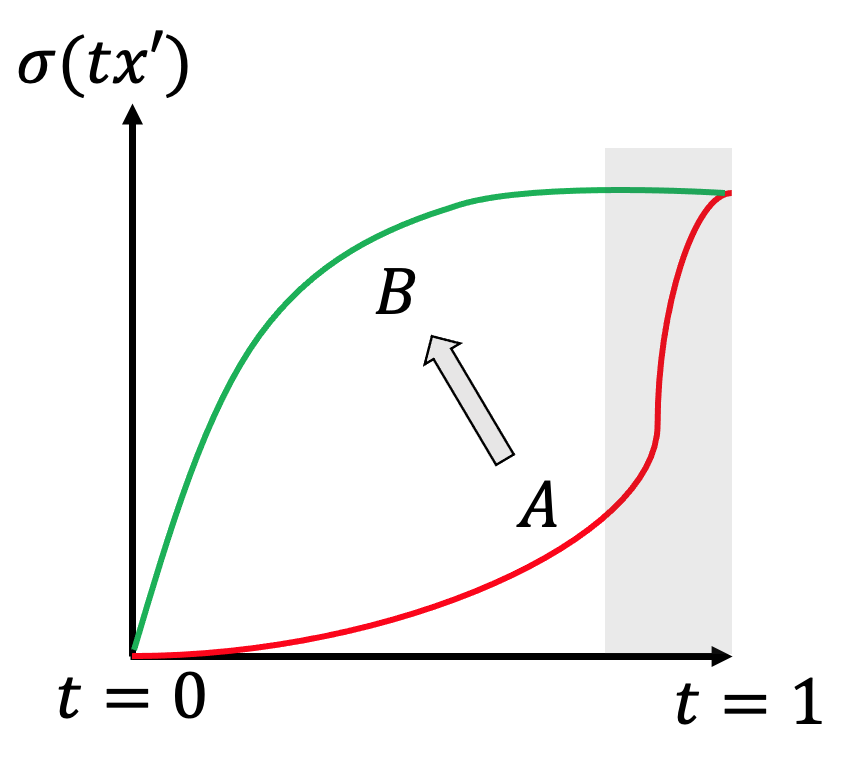}
        \caption{}
        \label{fig:illustration-curve}
     \end{subfigure}
    \caption{(a) A geometric view of the input space in a ReLU network. Dashed lines correspond to activation constraints while the colorful solid lines are piece-wise linear decision boundaries. Taking gradient of the model's output with respect to the input returns a vector that is orthogonal to a nearby boundary (points in the blue and green regions) or an extension of a nearby boundary (the point in the yellow region). (b) Curves of the model's sigmoid output $\sigma(t\xvec')$ against $t$.}
 \end{figure}

Note that for points in a given activation region $\mathcal{R}(p)$, the network $f$ can be expressed as a linear function, i.e. $\forall \xvec \in \mathcal{R}(p) . f(\xvec) = \mathbf{w}^\top_p\xvec + b_p$ where $\mathbf{w}_p$ is given by $\mathbf{w} = \partial f(\xvec) / \partial \xvec$~\citep{Jordan2019ProvableCF, DBLP:conf/nips/HaninR19}. 
Decision boundaries are thus piecewise-linear constraints, $f(\xvec) \geq 0$ for binary classifiers, or $f_i(\xvec) \geq f_j(\xvec)$ between classes $i$ and $j$ for a categorical classifier, with linear pieces corresponding to the activation region of $\xvec$. 
This leads us to the following: \emph{(1)} if a decision boundary crosses $\mathcal{R}(p)$, then $\mathbf{w}_p$ will be orthogonal to that boundary, and \emph{(2)} if a decision boundary does not cross the region $\mathcal{R}(p)$, then $\mathbf{w}_p$ is orthogonal to an \emph{extension} of a nearby boundary~\citep{fromherz20projections, Wang2021BoundaryAP}. 
In either case, the gradient with respect to the input captures information about a particular nearby decision boundary. Figure~\ref{fig:illustration-circle} summarizes this visually.

This analysis motivates the introduction of \emph{distributional influence} (Definition~\ref{def:distributional-influence}), which aggregates the gradients of the model at points in a given \emph{distribution of interest} (DoI) around $\xvec$.

\begin{definition}[Distributional Influence~\citep{influence-directed}]\label{def:distributional-influence}
    Given an input $\xvec$, a network $f: \mathbb{R}^d \rightarrow \mathbb{R}^m$, a class of interest $c$, and a distribution of interest $\mathcal{D}_\xvec$ which describes a reference neighborhood around $\xvec$, define the distributional influence as $\chi^c_{\mathcal{D}_\xvec}(\xvec) \defeq \E_{\xvec' \sim \mathcal{D}_\xvec} [\partial f_c(\xvec') / \partial \xvec']$. We write $S(\mathcal{D}_\xvec)$ to represent the support of $\mathcal{D}_\xvec$. When $m=1$, we write $\chi_{\mathcal{D}_\xvec}(\xvec) \defeq \E_{\xvec' \sim \mathcal{D}_\xvec} [\partial f(\xvec') / \partial \xvec']$.
\end{definition}

% Therefore, by aggregating the influence of points over a distribution of points around the counterfactual, called the \emph{distribution of interest} (DoI)~\citep{influence-directed}, we can capture the decision boundaries in a neighborhood of points defined by that DoI. 
% Following \citet{influence-directed}, we refer to this aggregated influence over the DoI as the \emph{distributional influence}. 
In \citet{influence-directed}, distributional influence is used to attribute the importance of a model's input and internal features on observed outcomes.
Following the connection between gradients and decision boundaries in ReLU networks, we leverage it to capture useful information about nearby decision boundaries as detailed in Section~\ref{sec:lipschitz}.
%The aggregation is over a support set $S(\mathcal{D})$ of DoI. We will discuss the choice DoI (and its support) for this paper in the next section.

% Thus, aggregating more gradients in a particular region allows us capture more decision boundaries at the same time. Following \citet{influence-directed}, 

% \citet{influence-directed} recently introduce a similar definition, \emph{distributional influence} (Def.~\ref{def:distributional-influence}), but is motivated to attribute the model's output to each input feature and aggregate the attributions over a distribution of interest (DoI) $\mathcal{D}$. Now, by the takeaways from the geometrical analysis of ReLU networks, we can interpret \emph{distributional influence} differently: distributional influence is a way of capturing the (aggregated) orthogonal vectors of nearby decision boundaries. The aggregation is over a support set $S(\mathcal{D})$ of DoI. We will discuss the choice DoI (and its support) for this paper in the next section.

% \subsection{Lipschitz Continuity and Consistency}

\subsection{Consistency and Continuity}
\label{sec:lipschitz}
Characterizing the precise effect such as random initialization have on the outcome of training is challenging. We approach this by modeling the differences that arise from small changes
such as a \emph{fine-tuning} of the original model, where the top layer of the model is re-trained and the parameters of rest of the layers are frozen. %motivated by the fact that a frozen backbone has been shown to be efficient in fine-tuning the many downstream tasks in practice~\citep{Devlin2019BERTPO, Jayram2019TransferLI, Kong2020PANNsLP}. 

We now introduce Theorem~\ref{theorem:loss-and-lipschitz}, which bounds the change on distributional influence when the model is fine-tuned at its top layer in terms of the model's Lipschitz continuity on the support of $\mathcal{D}_\xvec$.
This suggests that finding a high-confidence counterfactual example in a neighborhood with a lower Lipschitz constant may lead to lower invalidation after fine-tuning, given the relationship between nearby boundaries and influence described in the previous section.
%which shows that by finding a counterfactual with higher confidence, which exists in a neighborhood with a lower Lipschitz constant, one can decrease the change in a model's decision boundaries around that counterfactual over fine-tunings to the top layer of a model, thus increasing the stability of the counterfactual example.  
%which shows that the Lipschitz constant of a reference neighborhood around a counterfactual, together with the confidence of the model's output on the counterfactual, make up important components of the upper bound of the the difference in the distributional influence, i.e. the change in a decision boundaries, after a fine-tuning of the last layer.
%which shows that by finding a counterfactual with higher confidence, which exists in a neighborhood with a lower Lipschitz constant, one can decrease the change in a model's decision boundaries, thus increasing the stability of the counterfactual example.  

\newcommand{\wvec}{\ensuremath{\mathbf{w}}}
% \begin{theorem}\label{theorem:loss-and-lipschitz}
%     Let $f(\xvec) \defeq g(h(\xvec))$ be a ReLU network with a single logit output (i.e., a binary classifier) where $h(\xvec)$ is the output of the penultimate layer and $g(\xvec;\wvec) \defeq \wvec^\top \xvec$ is the top layer (we omit the bias as it does not contribute to the derivative of $g$). We denote $\sigma_\wvec = \sigma(g(\xvec;\wvec))$ as the sigmoid output at the point $\xvec$. Let $\mathcal{W} \defeq \{ \wvec' : ||\wvec - \wvec'|| \leq \Delta \}$ and $\chi_{\mathcal{D}_\xvec}(\xvec; \wvec)$ be the distributional influence w.r.t the distribution of interest $\mathcal{D}_\xvec$ when the top layer's weight is $\wvec$. If $h$ is $K$-Lipschitz in the support $S(\mathcal{D}_\xvec)$, the following inequality holds.
%     \begin{align}
%         \forall \wvec' \in \mathcal{W} .     ||\chi_{\mathcal{D}_\xvec}(\xvec; \wvec) - \chi_{\mathcal{D}_\xvec}(\xvec; \wvec')|| \leq K \left[\frac{\partial \sigma_\wvec}{\partial g} \cdot ||\wvec - \lambda\wvec'|| + C \right]
%     \end{align}
%     where $\lambda = \frac{\partial \sigma_\wvec}{\partial g} / \frac{\partial \sigma_{\wvec'}}{\partial g} $ and $C=\frac{1}{2}(||\wvec|| + \frac{1}{2}\Delta)$.
% \end{theorem}

\begin{theorem}\label{theorem:loss-and-lipschitz}
    Let $f(\xvec) \defeq \wvec^\top \cdot h(\xvec) + b$ be a ReLU network with a single logit output (i.e., a binary classifier), where $h(\xvec)$ is the output of the penultimate layer, and denote $\sigma_\wvec = \sigma(f(\xvec))$ as the sigmoid output of the model at $\xvec$.
    Let $\mathcal{W} \defeq \{ \wvec' : ||\wvec - \wvec'|| \leq \Delta \}$ and $\chi_{\mathcal{D}_\xvec}(\xvec; \wvec)$ be the distributional influence of $f$ when weights \wvec are used at the top layer. 
     If $h$ is $K$-Lipschitz in the support $S(\mathcal{D}_\xvec)$, the following inequality holds:
    \[
        \forall \wvec' \in \mathcal{W} .     ||\chi_{\mathcal{D}_\xvec}(\xvec; \wvec) - \chi_{\mathcal{D}_\xvec}(\xvec; \wvec')|| \leq K \left[\frac{\partial \sigma_\wvec}{\partial g} \cdot ||\wvec - \lambda\wvec'|| + C \right]
    \]
    where $\lambda = \frac{\partial \sigma_\wvec'}{\partial g} / \frac{\partial \sigma_{\wvec}}{\partial g} $ and $C=\frac{1}{2}(||\wvec|| + \frac{1}{2}\Delta)$.
\end{theorem}

\paragraph{Observations.} 
Theorem~\ref{theorem:loss-and-lipschitz} characterizes the extent to which a model's local decision boundaries, by proxy of influence, may change as a result of fine-tuning. 
This intuitively relates to the likelihood of a counterfactual's invalidation, as a point near a decision boundary undergoing a large shift is more likely to experience a change in prediction than one near a stable portion of the boundary.
As the two key ingredients in Theorem~\ref{theorem:loss-and-lipschitz} are the local Lipschitz constant and the model's confidence at $\xvec$, this suggests that searching for high-confidence points in neighborhoods with small Lipschitz constants will yield more consistent counterfactuals. 
While Theorem~\ref{theorem:loss-and-lipschitz} does not provide a direct bound on invalidation, and is limited to changes only at the network's top layer, we characterize the effectiveness of this heuristic in more general settings empirically in Section~\ref{sec:evaluation} after showing how to efficiently operationalize it in Section~\ref{sec:sns}.

\subsection{Finding Consistent Counterfactuals}
\label{sec:sns}
\label{RNS}
The results from the previous section suggest that counterfactuals with higher sigmoid output and lower Lipschitz Constants of the penultimate layer's output with respect to the DoI $\mathcal{D}_\xvec$ will be more consistent across related models. \emph{Stable Neighbor Search} (SNS) leverages this intuition to find consistent counterfactuals by searching for those with a low Lipschitz constant and confident counterfactual. We can find such points with
%This intuition is formalized by 
the objective in Equation~\ref{Eq:find_small_K}, which assumes a given counterfactual point $\xvec$. 
\begin{align}\label{Eq:find_small_K}
    \xvec_c = \arg\max_{\xvec' \in B(\xvec, \delta)} \left[\sigma(\xvec') - K_{S'} \right]
    \quad
    \text{such that}\ 
    F(\xvec_c; \theta) = F(\xvec; \theta)
\end{align} 
In Eq.~\ref{Eq:find_small_K} above and throughout this section, we assume that $F$ is a binary classifier with a single-logit output $f$, and sigmoid output $\sigma(f(\xvec))$. When $f$ is clear from the context, we directly write $\sigma(\xvec)$. The results are readily extended to multi-logit outputs by optimizing over the maximal logit at $\xvec$.
$K_{S'}$ is the Lipschitz Constant of the model's sigmoid output over the support $S(\mathcal{D}_{\xvec'})$. 
We relax the Lipschitz constant $K$ of the penultimate output in the Theorem~\ref{theorem:loss-and-lipschitz} to the Lipschitz constant of the entire network, as in practice any parameter in the network, and not just the top layer, may change. %Our empirical results in Sec~\ref{sec:evaluation} show that this relaxation is efficient to provide useful gradient information in the search for consistent counterfactuals. 

Leveraging a well-known relationship between the dual norm of the gradient and a function's Lipschitz constant~\citep{Paul}, we can rephrase this objective as shown in Equation~\ref{Eq:find_robust_neighbor}. 
Note that we assume $\ell_2$ norms throughout, so the dual remains $\ell_2$.
\begin{align}\label{Eq:find_robust_neighbor}
    \xvec_c = \arg\max_{\xvec' \in B(\xvec, \delta)} \left[\sigma(\xvec') - \max_{\myhat{\xvec} \in S(\mathcal{D}_{\xvec'})} ||\frac{\partial\sigma(\myhat{\xvec})}{\partial \myhat{\xvec}}|| \right]
    \quad
    \text{such that}\ 
    F(\xvec_c; \theta) = F(\xvec; \theta)
\end{align} 

\paragraph{Choice of DoI.} %By viewing gradients as normal vectors for decision boundaries, 
The choice of DoI determines the neighborhood of points from which we gain an understanding of the local decision boundary~\citep{Wang2021BoundaryAP}. 
%within which we project all points onto the nearby boundaries and aggregate the corresponding normal vectors~\citep{Wang2021BoundaryAP}. 
In this paper, following prior work, we choose $\mathcal{D}$ as $\text{Uniform}(\mathbf{0} \rightarrow \xvec)$, a uniform distribution over a linear path between a zero vector and the current input~\citep{sundararajan2017axiomatic}. That is, the set of points in $\mathcal{D}$ is $S(\mathcal{D}) \defeq \{t\xvec, t\in [0, 1]\}$.
%Our choice of DoI is motivated as follows: compared with other DoIs, i.e. a Gaussian or a Uniform distribution around the input~\citep{}, a Uniform distribution over a linear path can include more projections onto non-nearby boundaries of the input, which better captures the behavior of the global geometry of the model's output in the input space. When capturing more decision boundaries, the result of Theorem 2 ensures not only the nearby but also that further decision boundaries remain similar, which is expected to ensure our counterfactuals remain valid after the retraining. 
Equation~\ref{Eq:find_robust_neighbor-challenge-one} below updates the objective accordingly. 
\begin{align}\label{Eq:find_robust_neighbor-challenge-one}
    \xvec_c = \arg\max_{\xvec' \in B(\xvec, \delta)} \left[\sigma(\xvec') - \max_{t \in [0, 1]} ||\frac{ \partial \sigma(t\xvec')}{\partial (t\xvec')}|| \right]
    \quad
    \text{such that}\ 
    F(\xvec_c; \theta) = F(\xvec; \theta)
\end{align}

While Equation~(\ref{Eq:find_robust_neighbor-challenge-one}) provides an objective that uses only primitives that are readily available in most neural network frameworks, solving the inner objective using gradient descent requires second-order derivatives of the network, which is computationally prohibitive. In the following, we discuss a sequence of relaxations to Eq.~(\ref{Eq:find_robust_neighbor-challenge-one}) that provides resource-efficient objective function. 

\paragraph{Avoiding vacuous second-order derivatives.} %We first notice that 
There exists a lower-bound of the term $\max_{t \in [0, 1]} || \partial \sigma(t\xvec') / \partial (t\xvec')||$ by utilizing the following Proposition~\ref{prop:lowerbound-of-gradient-norm}, which allows us to relax Eq.~\ref{Eq:find_robust_neighbor-challenge-one} by maximizing a differentiable lower-bound of the gradient norm rather than the gradient norm itself.

% \paragraph{Avoiding vacuous second-order derivatives.} 
% To solve the first problem, we need a substitution for $||\nabla f(\myhat{\xvec}, \theta)||_2$ that will provide useful gradients for the outer minimization. 
% Ghorbani et al.~\cite{ghorbani2019interpretation} show that replacing ReLU activations with a related twice-differentiable activation, such as Softplus, can be a fruitful approach. 
% However, this can also lead to a model with a substantially different decision surface, which complicates enforcing the equality constraint over $F$ during optimization.
% Instead, we introduce a parameter $t$ over which we can take the derivative, in place of $\myhat{x}$, allowing the subsequent derivative with respect to $\xvec'$ to succeed.
% Proposition~\ref{theorem:lowerbound-of-gradient-norm} shows that such a $t$ exists, and how to construct it.

\begin{proposition}\label{prop:lowerbound-of-gradient-norm}
    Let $q$ be a differentiable, real-valued function in $\mathbb{R}^d$ and $S$ be the support set of $\text{Uniform}(\mathbf{0}\rightarrow\xvec)$. 
    Then for $\xvec' \in S$, 
    $
    ||\partial q(\xvec') / \partial \xvec'|| \ge ||\xvec||^{-1} |\partial q(r\xvec') / \partial r|_{r=1}|
    $.
\end{proposition}

% Proposition~\ref{theorem:lowerbound-of-gradient-norm} allows us to relax Eq.~\ref{Eq:find_robust_neighbor} by maximizing a differentiable lower-bound of the gradient norm rather than the gradient norm itself.
Noting that the constant factor $||\xvec||$ is irrelevant to the desired optimization problem, Equation~\ref{Eq:find_robust_neighbor-challenge-two} below updates the objective by fitting $\sigma$ into the place of $q$ in Proposition~\ref{prop:lowerbound-of-gradient-norm}. The absolute-value operator is omitted because the derivative of the sigmoid function is always non-negative. 
\begin{align}\label{Eq:find_robust_neighbor-challenge-two}
    \xvec_c = \arg\max_{\xvec' \in B(\xvec, \delta)} \left[\sigma(\xvec') - \max_{t \in [0, 1]} \frac{ \partial \sigma(t\xvec')}{\partial t} \right]
    \quad
    \text{such that}\ 
    F(\xvec_c; \theta) = F(\xvec; \theta)
\end{align}

The second term in Equation~\ref{Eq:find_robust_neighbor-challenge-two}, $- \max_{t \in [0, 1]}  \partial \sigma(t\xvec')/\partial t$, is interpreted by plotting the output score $\sigma(t\xvec')$ against the interpolation variable $t$ as shown in Fig.~\ref{fig:illustration-curve}. This term encourages finding a counterfactual point $\xvec_c$ where the outputs of the model for points between the zero vector $(t=1)$ and itself $(t=1)$ form a smooth and flattened curve B in Fig.~\ref{fig:illustration-curve}. Therefore, by incorporating the graph interpretation of $- \max_{t \in [0, 1]} \partial \sigma(t\xvec')/\partial t$ to find an solution of $\xvec_c$ that corresponds to curve B, we can instead try to increase the area under the curve of $\sigma(t\xvec')$ against $t$, which simplifies our objective function with replacing the inner-derivative with an integral shown in Equation~\ref{Eq:find_robust_neighbor-integral}.
\begin{align}\label{Eq:find_robust_neighbor-integral}
    \xvec_c = \arg\max_{\xvec' \in B(\xvec, \delta)} \left[\sigma(\xvec') + \int^1_0\sigma(t\xvec') dt  \right]
    \quad
    \text{such that}\ 
    F(\xvec_c; \theta) = F(\xvec; \theta)
\end{align}
One observation of the objective defined by Equation~\ref{Eq:find_robust_neighbor-integral} is that the first term $\sigma(\xvec')$ is redundant, as differentiating the second integral term already provides useful gradient information to increase $\sigma(\xvec')$. Equation~\ref{Eq:find_robust_neighbor-integral} thus yields our approach, \emph{Stable Neighbor Search}.

\begin{definition}[Stable Neighbor Search (SNS)]\label{def:SNS}
Given a starting counterfactual $\xvec$ for a network $F(\xvec)$, its \emph{stable neighbor} $\xvec_c$ of radius $\epsilon$ is the solution to the following objective: 
\[
\arg\max_{\xvec' \in B(\xvec, \delta)} \int^1_0\sigma(t\xvec') dt 
\]
\end{definition}

% We discuss the choice of $\epsilon$, which controls the size of the stable region around $\xvec_c$, in Sec.~\ref{sec:evaluation}.
To implement Definition~\ref{sec:sns}, the integral is replaced by a summation over a grid of points of a specified resolution, which controls the quality of the final approximation. 

% \subsection{Lipschitz Network}
% \input{lip_network}
% !TEX ROOT=./paper.tex
\section{Evaluation}\label{sec:evaluation}
In this section, we evaluate the extent of invalidation across five different counterfactual generation methods, including Stable Neighbor Search, over models trained with two sources of randomness in setup: \emph{1)} initial weights, and \emph{2)} leave-one-out differences in training data.
Our results show that Stable Neighbor Search consistently generates counterfactuals with lower invalidation rates than all other methods, in many cases eliminating invalidation altogether on tested points. 
Additionally, despite not explicitly minimizing cost, SNS counterfactuals manage to maintain low cost relative to other methods that aim to minimize invalidation.

\subsection{Setup}
\paragraph{Data.} Our experiments encompass several tabular classification datasets from the UCI database %(with MIT license)~\citep{uci}, 
including: German Credit, Taiwanese Credit-Default, Seizure, and Cardiotocography (CTG). 
We also include FICO HELOC~\citep{fico} and Warfarin Dosing~\citep{international2009estimation}. % Our use of HELOC complies with its Dataset Usage License~\citep{fico_license}.
All datasets have two classes except Warfarin, where we assume that the most favorable outcome (class 0) is the desired counterfactual for the other classes, and vice versa.
Further details of these datasets are included in Appendix~\ref{appendix:experiment-details-datasets}.

\paragraph{Baselines.} 
We compare SNS with the following baselines in terms of the invalidation rate. 
Further details about how we implement and configure these techniques are found in Appendix~\ref{appendix:experiment-detail-baselines}. \textbf{Min-Cost $\ell_1/\ell_2$}~\citep{wachter2018counterfactual}: we implement this by setting the appropriate parameters for the elastic-net loss~\citep{chen2018ead} in ART~\citep{art2018}. \textbf{Min-Cost $\epsilon$-PGD}~\citep{wachter2018counterfactual}: We perform Projected Gradient Descent (PGD) for an increasing sequence of $\epsilon$ until a counterfactual is found. \textbf{Pawelczyk et al.}~\citep{pmlr-v124-pawelczyk20a}: This method attempts to find counterfactual examples \emph{on the data manifold}, that are therefore more resistant to invalidation, by searching the latent space of a variational autoencoder, rather than the input space. \textbf{Looveren et al.}~\citep{van2019interpretable}: This method minimizes an elastic loss combined with a term that encourages finding examples on the data manifold.

%\begin{description}
% \item[Min-Cost $\ell_1/\ell_2$~\citep{wachter2018counterfactual}:] We implement this by setting the appropriate parameters for the elastic-net loss~\citep{chen2018ead} in ART~\citep{art2018}.
% \item[Min-$\epsilon$ PGD~\citep{madry2018towards}:] We perform Projected Gradient Descent (PGD) for an increasing sequence of $\epsilon$ until a counterfactual is found. 
% \item[Pawelczyk et al.~\citep{pmlr-v124-pawelczyk20a}:] This method attempts to find counterfactual examples \emph{on the data manifold}, that are therefore more resistent to invalidation, by searching the latent space of a variational autoencoder, rather than the input space.
% \item[Looveren et al.~\citep{van2019interpretable}:] This method minimizes an elastic loss combined with a term that encourages finding examples on the data manifold.
% \end{description}
We note that PGD was originally proposed in the context of adversarial adversarial examples~\citep{szegedy2013intriguing}. 
As has been noted in prior work, the problem of finding adversarial examples is mathematically identical to that of finding counterfactual examples~\citep{freiesleben2020counterfactual,browne2020semantics,sokol2019counterfactual,wachter2018counterfactual}. 
While solution sparsity is sometimes noted as a differentiator between the two, we note that techniques from both areas of research can be used with various $\ell_p$ metrics.
We measure cost in terms of both $\ell_1$ and $\ell_2$ norms, providing $\ell_2$ in the main body and $\ell_1$ in Appendix~\ref{appendix:ell1-results}.
% As the difference between these adversarial and counterfactual examples is an area of active research, we do not endeavor to solve this question.

\begin{table}[t]
    \resizebox{\textwidth}{!}{%
        \begin{tabular}{c|cc|cc|cc|cc|cc|cc}
            \multicolumn{13}{c}{\emph{Invalidation Rate}} \\
            \toprule 
            \textbf{Method}&\multicolumn{2}{c}{German Credit} & \multicolumn{2}{c}{Seizure} & \multicolumn{2}{c}{CTG} & \multicolumn{2}{c}{Warfarin}&\multicolumn{2}{c}{HELOC}& \multicolumn{2}{c}{Taiwanese Credit}\\
            \textbf&LOO & RS & LOO  & RS&LOO& RS&LOO & RS & LOO  & RS&LOO & RS \\
            \midrule
            Min. $\ell_1$ & 0.41 & 0.56 & - & - & 0.07 &  0.29& 0.44 & 0.35 & 0.30 &0.43 &0.30 & 0.78\\
            $\quad \quad$+SNS & 0.00 &0.07 & - & - & \textbf{0.00} & 0.01 & \textbf{0.00} & \textbf{0.00} & \textbf{0.00}& \textbf{0.00}& \textbf{0.00}&0.04 \\
            Min. $\ell_2$ &0.36 & 0.56 & 0.64 & 0.77 & 0.48 & 0.49& 0.35 &0.3 &0.55 & 0.61 & 0.27&0.72\\
            $\quad \quad$+SNS & 0.00 & \textbf{0.06} & \textbf{0.02} & 0.13 &\textbf{0.00} & \textbf{0.00}& \textbf{0.00} & \textbf{0.00} &\textbf{0.00}&\textbf{0.00}&\textbf{0.00}&\textbf{0.04}\\
            Min. $\epsilon$ PGD & 0.28 &  0.61 & 0.94 & 0.94& 0.04& 0.09 &0.10& 0.12& 0.04 & 0.11 &0.04 & 0.24\\
            $\quad \quad$+SNS & \textbf{0.00} & 0.12 &0.04 & 0.16 & \textbf{0.00} & \textbf{0.00} &0.01& 0.02& \textbf{0.00}&  \textbf{0.00} &\textbf{0.00} & 0.11\\
            Looveren et al. & 0.25 & 0.40 & 0.48 & 0.54 & 0.11&0.18 & 0.26 & 0.25 & 0.25 & 0.34& 0.29& 0.53\\
            Pawelczyk et al. & 0.20 &0.35 &0.16 &\textbf{0.11} &\textbf{0.00}&0.06 &0.02& 0.01 & 0.05& 0.15& 0.02 &0.20\\
            % \textsuperscript{15}N & \sfrac{1}{2} & 0.40 & 30.2 &&&\\
            % \rowcolor{black!5} \textsuperscript{16}O & 0 & 99.96 & -&& &\\
            % \textsuperscript{17}O & \sfrac{1}{2} & 0.04 & 40.4&&& \\
            \bottomrule
            \multicolumn{13}{c}{} \\
            \multicolumn{13}{c}{\emph{Counterfactual Cost ($\ell_2$)}} \\
            \toprule 
            \textbf{Method}&\multicolumn{2}{c}{German Credit} & \multicolumn{2}{c}{Seizure} & \multicolumn{2}{c}{CTG} & \multicolumn{2}{c}{Warfarin}&\multicolumn{2}{c}{HELOC}& \multicolumn{2}{c}{Taiwanese Credit}\\
            \midrule
            Min. $\ell_1$ & \multicolumn{2}{c}{1.33} & \multicolumn{2}{c}{-}& \multicolumn{2}{c}{0.17}&\multicolumn{2}{c}{0.50}&\multicolumn{2}{c}{0.24}& \multicolumn{2}{c}{1.56}\\
            Min. $\ell_2$ & \multicolumn{2}{c}{4.49} & \multicolumn{2}{c}{8.23}& \multicolumn{2}{c}{0.06}&\multicolumn{2}{c}{0.54}&\multicolumn{2}{c}{0.11}& \multicolumn{2}{c}{2.65}\\
            Looveren et al. & \multicolumn{2}{c}{5.37} & \multicolumn{2}{c}{8.40}& \multicolumn{2}{c}{0.11}&\multicolumn{2}{c}{1.03}&\multicolumn{2}{c}{0.45}& \multicolumn{2}{c}{2.82}\\
            Min. $\epsilon$ PGD & \multicolumn{2}{c}{1.02} & \multicolumn{2}{c}{1.36}& \multicolumn{2}{c}{0.08}&\multicolumn{2}{c}{0.31}&\multicolumn{2}{c}{0.32}& \multicolumn{2}{c}{0.75}\\
            % CW &  &  & && &\\
            % CW &  \multicolumn{2}{c}{0.37} & \multicolumn{2}{c}{2.52}& \multicolumn{2}{c}{-}&\multicolumn{2}{c}{-}&\multicolumn{2}{c}{0.19}& \multicolumn{2}{c}{0.77}\\
            \bottomrule
            % PGDs &  && && &\\
            Min.$\ell_1$ + SNS &\multicolumn{2}{c}{3.40} & \multicolumn{2}{c}{-}& \multicolumn{2}{c}{0.25}&\multicolumn{2}{c}{0.80}&\multicolumn{2}{c}{1.71}& \multicolumn{2}{c}{3.50}\\
            Min.$\ell_2$ + SNS &\multicolumn{2}{c}{6.23} & \multicolumn{2}{c}{9.60}& \multicolumn{2}{c}{\textbf{0.21}}&\multicolumn{2}{c}{0.90}&\multicolumn{2}{c}{1.71}& \multicolumn{2}{c}{4.68}\\
            PGD + SNS &\multicolumn{2}{c}{\textbf{3.03}} & \multicolumn{2}{c}{\textbf{3.60}}& \multicolumn{2}{c}{0.22}&\multicolumn{2}{c}{\textbf{0.50}}&\multicolumn{2}{c}{1.79}& \multicolumn{2}{c}{\textbf{2.78}}\\
            % CW &  &  & && &\\
            % CW+SNS & \multicolumn{2}{c}{\textbf{2.63}} & \multicolumn{2}{c}{\textbf{6.01}}& \multicolumn{2}{c}{-}&\multicolumn{2}{c}{-}&\multicolumn{2}{c}{1.75}& \multicolumn{2}{c}{3.37} \\
            Pawelczyk et al. & \multicolumn{2}{c}{7.15} & \multicolumn{2}{c}{13.66}& \multicolumn{2}{c}{1.07}&\multicolumn{2}{c}{2.62}&\multicolumn{2}{c}{\textbf{1.35}}& \multicolumn{2}{c}{4.24}\\
            \bottomrule
            \multicolumn{13}{c}{} \\
            \multicolumn{13}{c}{\emph{IV - Cost Correlation}} \\
            \toprule 
            \textbf{$R^2$} & \multicolumn{2}{c}{0.05} & \multicolumn{2}{c}{0.06}& \multicolumn{2}{c}{0.02}&\multicolumn{2}{c}{0.01}&\multicolumn{2}{c}{0.17}& \multicolumn{2}{c}{0.05}\\
            \multicolumn{13}{c}{}
        \end{tabular}%
        }
        \caption{The consistency of counterfactuals measured by invalidation rates. The average $\ell_2$ cost of different methods are also included. Results are aggregated over 100 networks for each experiment (RS and LOO). Lower invalidation rates and cost are more desirable. For $\ell_2$ cost, the best results are highlighted among three methods (separated by a line) with lower invalidation rates. If a method has significantly low success rate in generating counterfactual examples, we report `-'. In the last line, we present the $R^2$ correlation coefficient from a linear regression predicting invalidation percentage from cost. Small values indicate weak correlation.}
        \label{table_IV}
        \end{table}

\paragraph{Implementation of SNS.} SNS begins with a given counterfactual example as mentioned in Def.~\ref{def:SNS}, which we generate with Min. $\ell_1$/$\ell_2$ and Min. $\epsilon$ PGD. %The only hyper-parameter in SNS is $\alpha$ which is proportional to the radius of the local region over which we minimize the Lipschitz constant. 
We use the sum of 10 points to approximate the integral. %We find the resulting invalidation rate for SNS is not very sensitive to the choice of $\alpha$. We use $\alpha=0.5$ in all experiments and include the sensitivity analysis in Appendix~\ref{appendix:experiment-detail-sensitivity-of-alpha}. 

\paragraph{Retraining Controls.}
We prepare different models for the same dataset using Tensorflow 2.3.0 and all computations are done using a Titan RTX accelerator on a machine with 64 gigabytes of memory. 
We control the random seeds used by both numpy and Tensorflow, and enable deterministic GPU operations in Tensorflow~\citep{tf-determinism}. 
We evaluate the invalidation rate of counterfactual examples under changes in retraining stemming from the following two sources (see Appendix~\ref{appendix:experiment-detail-retraining} for more details on our training setup). \textbf{Leave-One-Out (LOO):} We select a random point (without replacement) to remove from the training data. Network parameters are initialized with the same values across runs. \textbf{Random Seed (RS):} Network parameters are initialized by incrementing the random seed across runs, while other hyperparameters and the data remain fixed.
% \begin{description}
    
%     \item[Leave-One-Out (LOO):] We select a random point (without replacement) to remove from the training data. Network parameters are initialized with the same values across runs.
    
%     \item[Random Seed (RS):] Network parameters are initialized by incrementing the random seed across runs, while other hyperparameters and the data remain fixed.

% \end{description}
We note that these sources of variation do not encompass the full set of sources that we are relevant to counterfactual invalidation, such as fine-tuning and changes in architecture or other hyperparameters.
However, they are straightforward to control, produce very similar models that nonetheless tend to invalidate counterfactuals, and they are not dependent on any deployment or data-specific considerations in the way that fine-tuning changes would be.
While we hope that our results are indicative of what might be observed across other sources, exploring invalidation in more depth in particular applications is important future work.

\paragraph{Metrics.}To benchmark the consistency of counterfactuals generated by different algorithms, we compute the mean invalidation rate (Def.~\ref{def:invalidation-rate}) over the validation split of each dataset.
To calculate the extent of correlation between cost and invalidation, as discussed in Section~\ref{sec:invalidation}, we perform a linear regression (\texttt{scipy.linregress}) between the costs for each valid counterfactual, across all five methods, with its invalidation rate across both LOO and RS differences. 
Table~\ref{table_IV} reports the resulting $R^2$ for each dataset.

\paragraph{Methodology.}
For each dataset, we train a ``base'' model and compute counterfactual examples using the five methods for each point in the validation split.
For each set of experiments (LOO or RS), we train $100$ additional models, and compute the invalidation rate between the base model and the $100$ variants. 
The results are shown in Table~\ref{table_IV}.

% For experiments investigating invalidation between models with a one-point difference in training set, we also set the numpy random seed, and use the same random initialization of model parameters across models. 
% For experiments evaluating invalidation over different random initialization, we set a new random seed for each model that is trained, and keep the same training set. 

%The German Credit, Seizure, and CTG models models have
% three hidden layers, of size 128, 64, and 16.  German Credit,
% Seizure, All models are trained for 100 epochs. German Credit models are trained
% with a batch size of 32,  Seizure ...
% The Lipschitz models are trained with the deel-lip package~\citep{} to ensure a specified Lipschitz constant throughout the entire model. They have the same architecture, batch size, training sets, and random seeds as the non-Lipschitz models. We train models with to have Lipschitz constant of 1, except when this deteriorates performance drastically, in which case we use a Lipschitz constant of 3 (CTG dataset).

\subsection{Results}

Looking at the invalidation results in Table~\ref{table_IV}, the most salient trend is apparent in the low invalidation rates of SNS compared to the other methods.
SNS achieves the lowest invalidation rate across 
%ten out of twelve experiments, and in one of these instances, there is just a one-point difference in the invalidation rate. 
all datasets in both LOO and RS experiments, except for on the Seizure dataset with RS variations, where there is a two-point difference in the invalidation rate. 
SNS generates counterfactuals with \emph{no} invalidation on CTG, Warfarin, and Heloc, and no invalidation over LOO differences on German Credit and Taiwanese Credit. 

Notably, this is down from invalidation rates as high as 61\% from other methods on Heloc, and $\approx10-50\%$ on others.
On Seizure, which had IV rates as high as 94\% from other methods, SNS achieves just 2\% (LOO) invalidation.
The closest competitor is the method of Pawelczyk et al.~\citep{pmlr-v124-pawelczyk20a}, which achieves zero invalidation in one case (CTG under LOO), but at significantly greater cost
--in five out of six cases, SNS produced less-costly counterfactuals, and in nearly every case the margin between the two is greater than $2\times$. %Note that costlier SNS counterfactuals are due to their high-cost starting points.%, as SNS searches for a nearby stable counterfactual to one given.
%--in every case SNS produced less-costly counterfactuals, and in nearly every case the margin between the two is greater than $2\times$.
%--in ten out of twelve experiments SNS produced less-costly counterfactuals, and in nearly every case the margin between the two is greater than $2\times$.

As discussed in Section~\ref{sec:invalidation}, while increasing cost is not a reliable way to generate stable counterfactuals for deep models, our results do show that stable counterfactuals tend to be more costly. 
The data suggests that greater-than-minimal cost appears to be necessary for stability. 
While SNS counterfactuals are much less costly than those generated by Pawelczyk et. al, they are consistently more costly than other methods that aim minimize cost without other constraints. 
To investigate the relationship between counterfactual cost and invalidation more closely, we report the $R^2$ coefficient of determination of a linear regression between the cost of each valid counterfactual generated and its invalidation rate in Table~\ref{table_IV}. Recall that a $R^2$ ranges from zero to one, with scores closer to zero indicating no linear relationship.
Notably, Table~\ref{table_IV} shows that the correlation between cost and invalidation is quite weak: the \emph{maximum} $R^2$ over all datasets is $0.17$ (Heloc), while most of the other datasets report coefficients that are much smaller--at or below 0.05. %As the $R^2$ coefficient is an indicator of the linear correlation between two variables, and ranges from zero to one, with zero indicating no linear relation, 

% \subsection{Cost-Invalidation Trade-off}

% \paragraph{Calculating Correlation between Cost and Invalidation.}
% We calculate the correlation between counterfactual cost and invalidation for each dataset by pooling the results of all counterfactual example methods together for each dataset, and using scipy.stats.lingress to calculate the least-squares regression between counterfactual cost and average invalidation rate over the $100$ models with either a leave-one-out difference in the training set, or trained with a different random seed, described earlier. We report the correlation through the r-value that the function retuSns directly.  

%We note that our results connecting instability to model loss point to an interesting intuition as to why counterfactuals from the data support are more stable: boundary changes \emph{outside} of the data distribution are much more unpredictable than boundary changes \emph{inside} the data distribution, as these changes do not affect the model's loss during training. Thus, data support counterfactuals may be more stable simply by virtue of the model's decision surface being more stable in areas around points in the data distribution, a phenomenon also noted by --et al~\citep{googlepaper}. 
%We note that SNS does not currently consider the feasibility of the counterfactuals it generates, nor does it consider whether the counterfactuals generated produce a realistic point. This may be made possible by limiting the search space of SNS, but we leave incorporating these desiderata as future work.
% !TEX ROOT=./paper.tex
\section{Related Work}\label{sec:related-work-2}
%This work focuses on evaluating and improving the consistency of prediction of counterfactual examples across duplicitous deep models, with inconsequential differences in the training set. 
% While counterfactual examples ...., 
% While counterfactuals enjoy popularity in the research literature~\citep{sokol2019counterfactual,wachter2018counterfactual,keane2020good,dandl2020multi,van2019interpretable,mahajan2019preserving,yang2020generating,verma2020counterfactual,pawelczyk2020learning,dhurandhar2018explanations,guidotti2018local}
Counterfactual examples enjoy popularity in the research literature~\citep{sokol2019counterfactual,wachter2018counterfactual,keane2020good,dandl2020multi,van2019interpretable,mahajan2019preserving,yang2020generating,verma2020counterfactual,pawelczyk2020learning,dhurandhar2018explanations,guidotti2018local}, especially in the wake of legislation increasing legal requirements on explanations of machine learning models~\citep{kaminski2019right,GDPR}.
However, recent work has pointed to problems with counterfactual examples that could occur during deployment~\citep{laugel2019issues,pmlr-v124-pawelczyk20a,barocas2020hidden,rawal2021i}. For example,~\citet{barocas2016big} point to the tension between the usefulness of a counterfactual and the ability to keep the explained model private. Previous work investigating the problem of invalidation, has pointed to cost as a heuristic for evaluating counterfactual invalidation at generation time~\citep{pmlr-v124-pawelczyk20a, rawal2021i}. We demonstrate that cost is not a reliable metric for predicting invalidation in \emph{deep} models, and show how the Lipschitz constant and confidence of a model around a counterfactual can be a more faithful guide to finding stable counterfactual examples.

While in this work, we address the problem of multiplicitious deep models producing varying outputs on \emph{counterfactual examples}, recent work has shown that there are large differences in model prediction behavior on ~\emph{any} input across small changes to the model~\citep{blackleave2021,marx2019,d2020underspecification}. Instability has also been shown to be a problem for %a broader class of explanation methods, such as 
gradient-based explanations, although this is largely studied in an adversarial context~\citep{dombrowski2019explanations,ghorbani2019interpretation,heo2019fooling}.

Within the related field of adversarial examples, there is a recent interest in \emph{adversarial transferability}~\citep{dong2018boosting,ilyas2019adversarialnotbugs,Xie_2019_CVPR}, where adversarial attacks are induced to transfer between models. In general, adversarial transferability concerns transferring attacks between extremely different models---e.g., trained on disjoint training sets. Meanwhile, in this work, we decrease counterfactual invalidation between very \emph{similar} models, in order to preserve recourse and explanation consistency. Interestingly,~\citet{advexamplesgoodfellow} suggest that transferability of adversarial examples is due to local linearity in deep networks. This supports our motivation: we find stable counterfactuals in more Lipschitz regions of the model, i.e. where it behaves (approximately) linearly. We note, however, that as linearity does not imply Lipschitzness, this insight does not provide a clear path to generating stable counterfactuals. We look forward to exploring the potential overlap between these two areas as future work.

\section{Conclusion}\label{sec:conclusion}

In this paper, we characterize the consistency of counterfactual examples in deep models, and demonstrate that counterfactual cost and consistency are not strongly correlated in deep models. To mitigate the problem of counterfactual instability in deep models, we introduce \emph{Stable Neighbor Search} (SNS), which finds stable counterfactual examples by leveraging the connection between the Lipschitz and confidence of the network around a counterfactual, and its consistency. 
% We demonstrate that SNS can generate more stable counterfactuals across changes in random seed or one-point differences in the training at a lower cost than several common counterfactual generation methods.
% While prior work points to increasing counterfactual cost as a method to increase stability, we demonstrate that this method is not always reliable in deep models---while higher cost is necessary for stable counterfactuals, it is not sufficient to ensure them. Instead, SNS leverages the theoretical observation that point with low local Lipschitz constant are harder for a model to change its prediction on during fine-tuning, as changing the model's prediction would incur a large loss. 
%We further demonstrate that SNS can generate more stable counterfactuals across changes in random seed or one-point differences in the training set over several benchmarking datasets over existing counterfactual searching methods.  
At a high level, our work adds to the growing perspective in the field of explainability that creating good explanations requires good models to begin with.

% In a broader sense, our results motivate the larger idea that high-quality model explanations will only come from sufficiently high-quality models: if a deep model has no regions with a low Lipschitz constant--i.e. its decision boundary is very steep everywhere, counterfactuals will be invalidated as a result of minuscule changes, as the model itself will be overly sensitive to these inconsequential changes. 

\section*{Acknowledgements}

This work was developed with the support of NSF grant CNS-1704845 as well as by DARPA and the Air Force Research Laboratory under agreement number FA8750-15-2-0277. The U.S. Government is authorized to reproduce and distribute reprints for Governmental purposes not with- standing any copyright notation thereon. The views, opinions, and/or findings expressed are those of the author(s) and should not be interpreted as representing the official views or policies of DARPA, the Air Force Research Lab- oratory, the National Science Foundation, or the U.S. Government.

% \clearpage
%!TEX root=./paper.tex

\section*{Ethics}

This paper demonstrates the problem of counterfactual invalidation in deep networks, and introduces a counterfactual generation method, Stable Neighbor Search (SNS), which creates counterfactual examples which yield consistent outcomes across nearby models. 
% As our focus in this paper is on the general properties of counterfactuals across domains, i.e. stability, we do not consider several aspects of counterfactual generation that require specific attention to the application domain--- for example, ensuring the counterfactuals are actionable, or respect the causal relationship of features in the application. %SNS can be used in conjunction with any other counterfactual generation method to find stable counterfactuals, and in cases where cost is of little concern, this may provide a partial solution.
% We look forward to incorporating these constraints into SNS in the context of a specific application domain as future work. 
We  note that the increased stability in counterfactual examples which SNS provides may eventually factor in to an engineer's, lawmaker's, or business' decision about what type of model to use: with the potential for more stable explanations, deep networks may seem more favorable. This, along with the ever-increasing zeal to incorporate neural networks in more applications, may lead practitioners to choose a deep model, when a simpler model may be a better fit for orthogonal reasons. %We urge model practitioners to consider the benefits and drawbacks of potential models choices before defaulting to the most accurate or complicated model. 
However, if used wisely, we believe SNS can lead to positive impacts, by lessening the invalidation of recourse to users who desire a different model outcome.

% \section*{Reproducability}

% We introduce novel theoretical results in this paper in Sections~\ref{sec:invalidation} and~\ref{sec:method}. Complete proofs for these results are can be found in Section~\ref{appendix:proofs} in the Appendix. We have included the code necessary to reproduce the results in this paper in the attached supplementary material. Specifically, we included files to train the relevant models, generate counterfactual examples of the different types presented, and perform analyses such as testing invalidation. The README file explains the necessary steps to download required packages, set up and environment, and how to run the scripts to reproduce the results.

%\newpage

\bibliography{iclr2022_conference}
\bibliographystyle{iclr2022_conference}

\appendix
\section{Proofs}\label{appendix:proofs}
\subsection{Theorem 1 and Lemma 1}
\noindent\textbf{Theorem 1} \textit{Suppose that $H_1, H_2$ are orthognal decision boundaries in a piecewise-linear network $F(\xvec) = sign\{w_1^\top ReLU(W_0\xvec)\}$, and let $\xvec$ be an arbitrary point in its domain. 
If the projections of $\xvec$ onto the corresponding halfspace constraints of $H_1, H_2$ are on $H_1$ and $H_2$, then there exists a point $\xvec'$ such that:
\begin{align*}
\mathit{1)}\ d(\xvec', H_2) = 0 & &
\mathit{2)}\ d(\xvec', H_2) < d(\xvec, H_2) & &
\mathit{3)}\ d(\xvec, H_1) \leq d(\xvec', H_1)
\end{align*}
where $d(\xvec, H_*)$ denotes the distance between $\xvec$ and the nearest point on a boundary $H_*$.}
\begin{proof}

    Let $u(\xvec)_i = W_0\xvec$ be the pre-activation of the neuron $i$-th output in the hidden layer. The status of the neuron therefore will have the following two status: \texttt{ON} if $u(\xvec)_i > 0$ and \texttt{OFF} otherwise. When a neuron is \texttt{ON}, the post-activation is identical to the pre-activation. Therefore, we can represent the ReLU function as a linear function of all neurons' activation status. Formally, the logit output of the network $F$ can be written as 
    \begin{align}
        f(\xvec) = w_1^\top \Lambda W_0\xvec
    \end{align} where $\Lambda$ is a diagonal matrix $diag([\lambda_0, \lambda_1, ..., \lambda_n])$ such that $\lambda_i = \mathbb{I}(u(\xvec)_i > 0)$. The network is a linear function within a neighborhood if all points in such a neighborhood have the same activation matrix $\Lambda$. For any two decision boundaries $H_1$ and $H_2$, the normal vectors of these decision boundaries can be written as $\nvec^\top_1 = w_1^\top \Lambda_1 W_0$ and $\nvec^\top_2 = w_1^\top \Lambda_2 W_0$, respectively, where $\Lambda_1$ and $\Lambda_2$ are determined by the activation status of internal neurons.

    For an input $\xvec$, if the projections of $\xvec$ onto the corresponding halfspace constraints of $H_1, H_2$ are on $H_1$ and $H_2$, then the distance $d(\xvec, H_1)$ and $d(\xvec, H_2)$ are given by projections as follows: 
    \begin{align}
        d(\xvec, H_1) = \frac{|\nvec^\top_1\xvec|}{||\nvec_1||_2}, \quad  d(\xvec, H_2) = \frac{|\nvec^\top_2\xvec|}{||\nvec_2||_2}
    \end{align}
    W.L.O.G. we assume $F(\xvec) = 1$  and $\nvec_1$ and $\nvec_2$ point towards $\xvec$. Let a point $\yvec$ defined as 
    \begin{align}
        \yvec &= \yvec' - \frac{|\nvec^\top_2\yvec'|\nvec_2}{||\nvec_2||^2_2}\\
        \yvec' &= \xvec + \eta\frac{\nvec_1}{||\nvec_1||_2}
    \end{align} where $\eta$ is tiny positive scalar such that $F(\yvec) = F(\xvec) =1$.
    We firstly show that $d(\yvec, H_2) = 0$ as follows: 
    \begin{align}
        d(\yvec, H_2) &= \frac{|\nvec_2^\top\yvec|}{||\nvec_2||_2}\\
                      &= \frac{|\nvec_2^\top(\yvec' - \frac{|\nvec^\top_2\yvec'|\nvec_2}{||\nvec_2||^2_2})|}{||\nvec_2||_2}\\
                      &= \frac{|\nvec^\top_2\yvec' - |\nvec^\top_2\yvec'||}{||\nvec_2||_2}\\
                      &= \frac{|\nvec^\top_2\yvec' - \nvec^\top_2\yvec'|}{||\nvec_2||_2} \quad (\eta \text{ is tiny so } \nvec_2\text{ points to }\yvec')\\ 
                      &= 0
    \end{align} We secondly show that $d(\yvec, H_1)>d(\xvec, H_1)$ as follows: 
    \begin{align}
        d(\yvec, H_1) &= \frac{|\nvec_1^\top \yvec|}{||\nvec_1||_2}\\
                      &=\frac{|\nvec_1^\top (\yvec' - \frac{|\nvec^\top_2\yvec'|\nvec_2}{||\nvec_2||^2_2})|}{||\nvec_1||_2}\\
                      &= \frac{|\nvec_1^\top (\xvec + \eta\frac{\nvec_1}{||\nvec_1||_2}- \frac{|\nvec^\top_2(\xvec + \eta\frac{\nvec_1}{||\nvec_1||_2})|\nvec_2}{||\nvec_2||^2_2})|}{||\nvec_1||_2}\\
                      &= \frac{|\nvec_1^\top\xvec + \eta||\nvec_1||_2 - \nvec_1^\top\nvec_2\frac{|\nvec^\top_2(\xvec + \eta\frac{\nvec_1}{||\nvec_1||_2})|}{||\nvec_2||^2_2})|}{||\nvec_1||_2}\\
                      &= \frac{|\nvec_1^\top\xvec + \eta||\nvec_1||_2|}{||\nvec_1||_2} \quad (\text{$H_1$ and $H_2$ are orthogonal})\\  
                      &\geq \frac{|\nvec_1^\top\xvec|}{||\nvec_1||_2} = d(\xvec, H_1)
    \end{align}
    The proof of Theorem 1 is complete. 
    \end{proof}

    \noindent\textbf{Lemma 1} \textit{Let $H_1, H_2, F$ and $\xvec$ be defined as in Theorem 1. 
    If the projections of $\xvec$ onto the corresponding halfspace constraints of $H_1, H_2$ are on $H_1$ and $H_2$, but there \emph{does not} exist a point $\xvec'$ satisfying \emph{(2)} and \emph{(3)} from Theorem 1, then $H_1$ = $H_2$.}

    Note if we remove the assumption that $H_1$ and $H_2$ are orthogonal, we will show that Theorem 1 will hold by condition. Let $m(
        \xvec
    ) = |\nvec_1^\top\xvec + \eta||\nvec_1||_2 - \nvec_1^\top\nvec_2\frac{|\nvec^\top_2(\xvec + \eta\frac{\nvec_1}{||\nvec_1||_2})|}{||\nvec_2||^2_2})|$. Assume the angle between the normal vectors of $H_1$ and $H_2$ is $\theta$ such that $\nvec^\top_1 \nvec_2 = ||\nvec_1||_2 ||\nvec_2||_2 \cos\theta$. 
    \begin{align}
        m(\xvec) &=  |\nvec_1^\top\xvec + \eta||\nvec_1||_2 - \nvec_1^\top\nvec_2\frac{|\nvec^\top_2(\xvec + \eta\frac{\nvec_1}{||\nvec_1||_2})|}{||\nvec_2||^2_2})| \\
         &=|\nvec_1^\top\xvec + \eta(||\nvec_1||_2- \frac{\nvec^\top_1\nvec_2\cdot\nvec^\top_2\nvec_1}{||\nvec_2||^2_2||\nvec_1||_2}) - \frac{\nvec^\top_1\nvec_2\cdot\nvec^\top_2\xvec}{||\nvec_2||^2_2}|\\
         &= |\nvec_1^\top\xvec + \eta(1-\cos^2\theta)||\nvec_1||_2 - \nvec_1\xvec\cos\theta|
    \end{align} Since $d(\yvec, H_1) \propto m(\xvec)$ and $d(\xvec, H_1) \propto |\nvec^\top_1\xvec|$ and they share they same denominator $||\nvec_1||_2$. In order to have $m(\xvec)>|\nvec^\top_1\xvec|$, we just need $\eta(1-\cos^2\theta)||\nvec_1||_2 - \nvec_1\xvec\cos\theta > 0$, which means we need to find a $\eta$ such that $\eta(1-\cos^2\theta)||\nvec_1||_2 > \nvec_1\xvec\cos\theta$. Moving terms around we have the following inequality:
    \begin{align}
        \eta > \frac{\nvec_1\xvec\cos\theta}{(1-\cos^2\theta)||\nvec_1||_2} = \frac{||\xvec||_2}{\frac{1}{\cos\theta}-\cos\theta}
    \end{align}
    The RHS goes to 0 when $\theta \rightarrow \frac{\pi}{2}$, which corresponds to the situation of Theorem 1. When $\theta \rightarrow 0$ ($H_1 = H_2$), RHS goes to $\infty$, which means we cannot find a point $\yvec$ satisfying the Theorem 1, which completes the proof of Lemma 1.

    \subsection{Theorem 2 and Proposition 1}

% \noindent\textbf{Theorem 2}\textit{
%     Let $F(\xvec;\theta^{(i)}) \defeq g(h(\xvec))$ be a ReLU network with a single logit output (i.e., a binary classifier) where $h(x)$ is the output of the penultimate layer and $g(x;\wvec) \defeq \wvec^\top \xvec$ is the top layer (we omit the bias as it does not contribute to the derivative of $g$). We denote $\sigma(\xvec;\wvec)$ as the sigmoid output at the point $\xvec$. Let $\mathcal{W} \defeq \{ w' |||\wvec - \wvec'|| \leq \Delta \}$ and $\chi_\mathcal{D}(\xvec; \wvec)$ be the distributional influence with respect to the distribution of interest $\mathcal{D}$ when the top layer's weight is $w$. If $h$ is $(K, \mathcal{D})$-distrbutional Lipschitz, the following inequality holds:
%     \begin{align}
%         \forall \wvec' \in \mathcal{W},     ||\chi_\mathcal{D}(\xvec; \wvec) - \chi_\mathcal{D}(\xvec; \wvec')|| \leq K \left[d\sigma(\xvec,\wvec)||\wvec - \lambda\wvec'|| + C \right]
%     \end{align}
%     where $d\sigma(\xvec;\wvec) = \sigma(\xvec;\wvec)(1-\sigma(\xvec;\wvec)) $,  $\lambda = d\sigma(\xvec;\wvec) / d\sigma(\xvec;\wvec') $ and $C=\frac{1}{2}(||w|| + \frac{1}{2}\Delta)$.}
\noindent\textbf{Theorem 2}\textit{
    Let $f(\xvec) \defeq \wvec^\top \cdot h(\xvec) + b$ be a ReLU network with a single logit output (i.e., a binary classifier), where $h(\xvec)$ is the output of the penultimate layer, and denote $\sigma_\wvec = \sigma(f(\xvec))$ as the sigmoid output of the model at $\xvec$.
    Let $\mathcal{W} \defeq \{ \wvec' : ||\wvec - \wvec'|| \leq \Delta \}$ and $\chi_{\mathcal{D}_\xvec}(\xvec; \wvec)$ be the distributional influence of $f$ when weights \wvec are used at the top layer. 
     If $h$ is $K$-Lipschitz in the support $S(\mathcal{D}_\xvec)$, the following inequality holds:
    \[
        \forall \wvec' \in \mathcal{W} .     ||\chi_{\mathcal{D}_\xvec}(\xvec; \wvec) - \chi_{\mathcal{D}_\xvec}(\xvec; \wvec')|| \leq K \left[\frac{\partial \sigma_\wvec}{\partial g} \cdot ||\wvec - \lambda\wvec'|| + C \right]
    \]
    where $\lambda = \frac{\partial \sigma_\wvec}{\partial g} / \frac{\partial \sigma_{\wvec'}}{\partial g} $ and $C=\frac{1}{2}(||\wvec|| + \frac{1}{2}\Delta)$.
}

\begin{proof}
    Consider a ReLU network as $g(h(\xvec))$. We first write out the expression of $h(x)$:
    \begin{align}
        h(x)  = \phi_{N-1}(W_{N-1}(\cdots \phi_1(W_1 x + b_1)) + b_{N-2})
    \end{align}
    where $W_i, b_i$ are the parameters for the $i$-th layer and $\phi_i(\cdot)$ is the corresponding ReLU activation. By the definition of the distributional influence, 
    \begin{align}
        \chi_\mathcal{D}(\xvec; \wvec) &= \mathbb{E}_{\zvec \sim \mathcal{D}(\xvec)}{\frac{\partial \sigma(g(h(\zvec);\wvec))}{\partial \zvec}}\\
        &= \mathbb{E}_{\zvec \sim \mathcal{D}(\xvec)} \frac{\sigma(g)}{\partial g} \frac{\partial g(h;\wvec)}{\partial h} \frac{\partial h(\zvec)}{\partial \zvec}\\
        &= \mathbb{E}_{\zvec \sim \mathcal{D}(\xvec)} \left[\sigma(\zvec;\wvec)(1-\sigma(\zvec;\wvec)) \wvec \prod^{N-1}_{i=1} (W_{i} \Lambda_i(\zvec))^\top\right]\\
    \end{align}
    where $W_{i}$ is the weight of the layer $l_i$ if $l_i$ is a dense layer or the equivalent weight matrix of a convolutional layer and $\Lambda_{i}(\zvec)$ is an diagonal matrix with each diagonal entry being 1 if the neuron is activated or 0 other wise when evaluated at the point $\zvec$. 
    \begin{align}
    ||\chi_\mathcal{D}(\xvec; \wvec) - \chi_\mathcal{D}(\xvec; \wvec')|| &= ||\mathbb{E}_{\zvec \sim \mathcal{D}(\xvec)} \left[\sigma(\zvec;\wvec)(1-\sigma(\zvec;\wvec)) \wvec \prod^{N-1}_{i=1} (W_{i} \Lambda_i(\zvec))^\top\right] \\&- \mathbb{E}_{\zvec \sim \mathcal{D}(\xvec)} \left[\sigma(\zvec;\wvec')(1-\sigma(\zvec;\wvec')) \wvec' \prod^{N-1}_{i=1} (W_{i} \Lambda_i(\zvec))^\top\right] ||\\
    &= ||\mathbb{E}_{\zvec \sim \mathcal{D}(\xvec)}\left[(\sigma(\zvec;\wvec)(1-\sigma(\zvec;\wvec)) \wvec - \sigma(\zvec;\wvec')(1-\sigma(\zvec;\wvec')) \wvec') \prod^{N-1}_{i=1} (W_{i} \Lambda_i(\zvec))^\top\right]||\\
    &\leq \mathbb{E}_{\zvec \sim \mathcal{D}(\xvec)} \left[||(\sigma(\zvec;\wvec)(1-\sigma(\zvec;\wvec)) \wvec - \sigma(\zvec;\wvec')(1-\sigma(\zvec;\wvec')) \wvec') \prod^{N-1}_{i=1} (W_{i} \Lambda_i(\zvec))^\top||\right]\\
    &(\text{Jensen's Inequality})\\
    &\leq \mathbb{E}_{\zvec \sim \mathcal{D}(\xvec)} \left[||(\sigma(\zvec;\wvec)(1-\sigma(\zvec;\wvec)) \wvec - \sigma(\zvec;\wvec')(1-\sigma(\zvec;\wvec')) \wvec')||\cdot|| \prod^{N-1}_{i=1} (W_{i} \Lambda_i(\zvec))^\top||\right]\\
    &(\text{By the definition of matrix operator norm})\\
    % &\leq ||\mathbb{E}_{\zvec \sim \mathcal{D}(\xvec)}\left[(\sigma(\zvec;\wvec)(1-\sigma(\zvec;\wvec)) \wvec - \sigma(\zvec;\wvec')(1-\sigma(\zvec;\wvec')) \wvec')\right]|| \\&\cdot ||\mathbb{E}_{\zvec \sim \mathcal{D}(\xvec)}\left[\prod^{N-1}_{i=1} (W_{i} \Lambda_i(\zvec))^\top\right]|| \\
    &\leq \mathbb{E}_{\zvec \sim \mathcal{D} (\xvec)}||\left[(\sigma(\zvec;\wvec)(1-\sigma(\zvec;\wvec)) \wvec - \sigma(\zvec;\wvec')(1-\sigma(\zvec;\wvec')) \wvec')\right]|| \\&\cdot \mathbb{E}_{\zvec \sim \mathcal{D}(\xvec)}||\left[\prod^{N-1}_{i=1} (W_{i} \Lambda_i(\zvec))^\top\right]|| 
    \end{align}
    We simplify the notation by defining $$d\sigma(\zvec;\wvec') \defeq \sigma(\zvec;\wvec')(1-\sigma(\zvec;\wvec')$$ and we denote $d\sigma(\zvec;\wvec) = d\sigma(\xvec;\wvec) +\delta(\zvec;\wvec)$ and $d\sigma(\zvec;\wvec') = d\sigma(\xvec;\wvec') +\delta(\zvec;\wvec')$. Note that $\delta \leq \frac{1}{4}$ because $d\sigma \in [0, \frac{1}{4}]$. Therefore, the first term can be simplified as
    \begin{align}
        \mathbb{E}_{\zvec \sim \mathcal{D} (\xvec)}&||(\sigma(\zvec;\wvec)(1-\sigma(\zvec;\wvec)) \wvec - \sigma(\zvec;\wvec')(1-\sigma(\zvec;\wvec')) \wvec')||\\ &= \mathbb{E}_{\zvec \sim \mathcal{D} (\xvec)}|| d\sigma(\xvec,\wvec)\wvec - d\sigma(\xvec,\wvec')\wvec' + \delta(\zvec;\wvec)\wvec  - \delta(\zvec;\wvec')\wvec'  ||\\
        & \leq ||d\sigma(\xvec,\wvec)\wvec - d\sigma(\xvec,\wvec')\wvec'|| + \mathbb{E}_{\zvec \sim \mathcal{D}}||\delta(\zvec;\wvec)\wvec  - \delta(\zvec;\wvec')\wvec' || \quad \text{(Triangle Inequality)} \\
        &\leq ||d\sigma(\xvec,\wvec)\wvec - d\sigma(\xvec,\wvec')\wvec'|| + \mathbb{E}_{\zvec \sim \mathcal{D}}||\delta(\zvec;\wvec)\wvec|| + \mathbb{E}_{\zvec \sim \mathcal{D}}||\delta(\zvec;\wvec')\wvec' || \\
        &\leq ||d\sigma(\xvec,\wvec)\wvec - d\sigma(\xvec,\wvec')\wvec'|| + \mathbb{E}_{\zvec \sim \mathcal{D}}|\delta(\zvec;\wvec)|||\wvec|| + \mathbb{E}_{\zvec \sim \mathcal{D}}|\delta(\zvec;\wvec')|||\wvec' ||\\
        &\leq ||d\sigma(\xvec,\wvec)\wvec - d\sigma(\xvec,\wvec')\wvec'|| + \frac{1}{4}(||w|| + ||w'||) \qquad (\delta \leq \frac{1}{4}) \\
        &\leq ||d\sigma(\xvec,\wvec)\wvec - d\sigma(\xvec,\wvec')\wvec'|| + \frac{1}{2}(||w|| + \frac{1}{2}\Delta)
    \end{align}
    Let $$\lambda \defeq \frac{d\sigma(\xvec,\wvec')}{d\sigma(\xvec,\wvec)}$$
    We have
    \begin{align}
         \mathbb{E}_{\zvec \sim \mathcal{D} (\xvec)}&||(\sigma(\zvec;\wvec)(1-\sigma(\zvec;\wvec)) \wvec - \sigma(\zvec;\wvec')(1-\sigma(\zvec;\wvec')) \wvec')|| \leq d\sigma(\xvec,\wvec) ||\wvec - \lambda\wvec'|| + \frac{1}{2}(||w|| + \frac{1}{2}\Delta)
    \end{align}
    Now we take a look at the second part
    \begin{align}
        \mathbb{E}_{\zvec \sim \mathcal{D}(\xvec)}||\left[\prod^{N-1}_{i=1} (W_{i} \Lambda_i(\zvec))^\top\right]|| \leq \sup_{\zvec \sim \mathcal{D}(\xvec)} ||\left[\prod^{N-1}_{i=1} (W_{i} \Lambda_i(\zvec))^\top\right]||\\
    \end{align}
    The RHS part of the inequality is a direct consequence of the definition of the Lipschitz Constant. Therefore, we have 
    \begin{align}
        \mathbb{E}_{\zvec \sim \mathcal{D}(\xvec)}||\left[\prod^{N-1}_{i=1} (W_{i} \Lambda_i(\zvec))^\top\right]|| \leq K\\
    \end{align}
\end{proof}

Put together,  we show that:
\begin{align}
    ||\chi_\mathcal{D}(\xvec; \wvec) - \chi_\mathcal{D}(\xvec; \wvec')|| \leq K \left[d\sigma(\xvec,\wvec)||\wvec - \lambda\wvec'|| + \frac{1}{2}(||w|| + \frac{1}{2}\Delta) \right]
\end{align}
By denoting $C = \frac{1}{2}(||w|| + \frac{1}{2}\Delta)$, we finish our proof.

    \subsection{Proposition 1}

    \noindent\textbf{Proposition 1}\textit{
    Let $q$ be a differentiable, real-valued function in $\mathbb{R}^d$ and $S$ be the support set of $\text{Uniform}(\mathbf{0}\rightarrow\xvec)$. $\forall\xvec' \in S$,
    \[
    ||\frac{\partial q(\xvec')}{\partial \xvec'}|| \ge ||\xvec||^{-1} |\frac{\partial q(r\xvec')}{\partial r}|_{r=1}|
    \]}
    \begin{proof}
        First, we show that $\forall \xvec' \in S$
        \begin{align}
            |\frac{\partial q(\xvec')}{\partial \xvec'}^\top\cdot\xvec'| \leq ||\frac{\partial q(\xvec')}{\partial \xvec'}||\cdot ||\xvec'|| \quad (\text{Cauchy–Schwarz})
        \end{align} By the construction of $\xvec'$ we know $||\xvec'|| \leq ||\xvec||$; therefore, 
        \begin{align}
            |\frac{\partial q(\xvec')}{\partial \xvec'}^\top\cdot\xvec'| &\leq ||\frac{\partial q(\xvec')}{\partial \xvec'}|| \cdot ||\xvec||\\
            ||\frac{\partial q(\xvec')}{\partial \xvec'}|| & \geq ||\xvec||^{-1} |\frac{\partial q(\xvec')}{\partial \xvec'}^\top\cdot\xvec'|
        \end{align}
        Now consider a function $p(r;\xvec') = r\xvec'$. Then we show a trick of chain rule.
        \begin{align}
            \frac{\partial q(p)}{\partial r} = \frac{\partial q(p)}{p}^\top \cdot \frac{\partial p(r;\xvec')}{\partial r} = \frac{\partial q(p)}{p}^\top \cdot \xvec'
        \end{align}
         Replacing the notation $p$ with $\xvec'$ in $\frac{\partial q(\xvec')}{\partial \xvec'}^\top$ does not change the computation of taking the Jacobian of $q$'s output with respect to the input; therefore, we show that
         
         \begin{align}
             \frac{\partial q(\xvec')}{\partial \xvec'}^\top\cdot\xvec' =\frac{\partial q(p)}{p}^\top|_{r=1} \cdot \xvec' = \frac{\partial q(p)}{\partial r}|_{r=1} = \frac{\partial q(r\xvec')}{\partial r}|_{r=1}
         \end{align}
         
        We therefore complete the proof of Proposition 1 by showing
        \begin{align}
            ||\frac{\partial q(\xvec')}{\partial \xvec'}|| \geq ||\xvec||^{-1} |\frac{\partial q(r\xvec')}{\partial r}|_{r=1}|
        \end{align}
    \end{proof}

\section{Experiment Details}\label{appendix:experiment-details}

\subsection{Meta information for Datasets}\label{appendix:experiment-details-datasets} 

The German Credit~\cite{uci} and Taiwanese Credit~\cite{uci} data sets consist of individuals financial data, with a binary response indicating their creditworthiness. For the German Credit~\cite{uci} dataset, there are 1000 points, and 20 attributes. We one-hot encode the data to get 61 features, and standardize the data to zero mean and unit variance using SKLearn Standard scaler. We partitioned the data intro a training set of 700 and a test set of 200.  
The Taiwanese Credit~\cite{uci} dataset has 30,000 instances with 24 attributes. We one-hot encode the data to get 32 features and normalize the data to be between zero and one. We partitioned the data intro a training set of 22500 and a test set of 7500.  

The HELOC dataset~\cite{fico} contains anonymized information about the Home Equity Line of Credit applications by homeowners in the US, with a binary response indicating whether or not the applicant has even been more than 90 days delinquent for a payment. The dataset consists of 10459 rows and 23 features, some of which we one-hot encode to get a dataset of 10459 rows and 40 features. We normalize all features to be between zero and one, and create a train split of 7,844 and a validation split of 2,615.

The Seizure~\cite{uci} dataset comprises time-series EEG recordings for 500 individuals, with a binary response indicating the occurrence of a seizure. This is represented as 11500 rows with 178 features each. We split this into 7,950 train points and 3,550 test points. We standardize the numeric features to zero mean and unit variance. 

The CTG~\cite{uci} dataset comprises of 2126 fetal cardiotocograms processed and labeled by expert obstetricians into three classes of fetuses, healthy, suspect, and pathological.  We have turned this into a binary response between healthy and other classes. We split the data into 1,700 train points and a validation split of 425. There are 21 features for each instance, which we normalize to be between zero and one. 

The Warfain dataset is collected by the International Warfarin
Pharmacogenetics Consortium~\cite{international2009estimation} about patients who were
prescribed warfarin. We removed rows with missing
values, 4819 patients remained in the dataset. The inputs to the
model are demographic (age, height, weight, race), medical
(use of amiodarone, use of enzyme inducer), and genetic
(VKORC1, CYP2C9) attributes. Age, height, and weight are
real-valued and were scaled to zero mean and unit variance.
The medical attributes take binary values, and the remaining
attributes were one-hot encoded. The output is the weekly dose
of warfarin in milligrams, which we encode as "low", "medium", or "high", following~\cite{international2009estimation}.

The UCI datasets are under an MIT license, and Warfarin datasets are under a Creative Commons License.~\cite{uci,international2009estimation}. The license for the FICO HELOC dataset is available at the dataset challenge website, and allows use for research purposes~\cite{fico_license}.

%The German Credit and Taiwanese Credit data sets consists of individuals’ financial data, with a binary response indicating their creditworthiness. The HELOC dataset contains anaoymized information about the Home Equity Line of Credit applications by homeowners in the US, with a binary response indicating whether or not the applicant has even been more than 90 days delinquent for a payment. The Seizure dataset comprises time-series EEG recordings for 500 individuals, with a binary response indicating the occurrence of a seizure. The Warfarin Dosing dataset consists of indivduals health and demographic information, with warfarin dosage recommendations of "low", "medium," and "high" as labels. The CTG dataset comprises of 2126 fetal cardiotocograms processed and labeled by expert obstetricians into three classes of fetuses, healthy, suspect, and pathological. We have turned this into a binary response between healthy and other classes. Further information about these datasets and the preprocessing steps we apply can be found in the supplementary material. 

\subsection{Hyper-parameters and Model Architectures }\label{appendix:experiment-details-hyperparams} 
The German Credit and Seizure models have
three hidden layers, of size 128, 64, and 16. Models on the Taiwanese dataset have two hidden layers of 32 and 16, and models on the HELOC dataset have two deep layers with sizes 100 and 32. The Warfarin models have one hidden layer of 100. The CTG models have three layers, of sizes 100, 32, and 16. 
German Credit,
Adult, Seizure, Taiwanese, CTG and Warfarin models are trained for 100 epochs; HELOC models are trained for 50 epochs. German Credit models are trained
with a batch size of 32; Adult, Seizure, and Warfarin models with
batch sizes of 128; Taiwanese Credit models with batch sizes of 512, and CTG models with a batch size of 16. All models are trained with keras' Adam optimizer with the default parameters.

\subsection{Implementation of Baseline Methods}\label{appendix:experiment-detail-baselines}
We describe the parameters specific to each baseline method here. Common choices of hyper-parameters are shown in Table~\ref{table_param}.
\begin{description}
    \item[Min-Cost $\ell_1/\ell_2$]\cite{wachter2018counterfactual} We implement this by setting $\beta=1.0$ for $\ell_1$ (or $\beta=0.0$ for $\ell_2$) and \texttt{confidence}=$0.5$ for the elastic-net loss~\cite{chen2018ead} in ART~\cite{art2018}. 
    \item[Min-$\epsilon$ PGD~\cite{madry2018towards}:] For a given $\epsilon$, we use 10 interpolations between 0 and the current $\epsilon$ as the norm bound in each PGD attack. The step size is set to $2*\epsilon_c/$ \texttt{max\_steps} where $\epsilon_c$ is the norm bound used. The maximum allowed norm bound is the median of the $\ell_2$ norm of data points in the training set. 
    \item[Pawelczyk et al.~\cite{pmlr-v124-pawelczyk20a}:] We train an AutoEncoder (AE) instead of a Variational AutoEncoder (VAE) to estimate the data manifold. Given that VAE jointly estimate the mean and the standard deviation of the latent distribution, it creates non-deterministic latent representation for the same input. In the contact with Pawelczyk et al., we are informed that we can only use the mean as the latent representation for an input; therefore, by taking out the standard deviation from a VAE, we instead train a AE that produces deterministic latent representation for each input. When searching for the latent representation of a counterfactual, we use random search as proposed by Pawelczyk et al.~\cite{pmlr-v124-pawelczyk20a}: we randomly sample 1280 points around the latet representation of an input within a norm bound of $1.0$ in the latent space. When generating random points, we use a fixed random seed 2021. If there are multiple counterfactuals, we return the one that is closest to the input. For all datasets, we use the following architecture for the hidden layers: 1024-128-32-128-1024.
    \item[Looveren et al.~\cite{van2019interpretable}:] We use the public implementation of this method\footnote{\url{https://docs.seldon.io/projects/alibi/en/stable/methods/CFProto.html}}. We use k-d trees with $k=20$ to estimate the data manifold as the curre implementation only supports an AE where the input features must be between 0 and 1, while our dataset are not normalized into this range. The rest of the hyper-parameters are default values from the implementation: \texttt{theta}=100, \texttt{max\_iterations}=100. This implementation only supports for non-eager mode so we turn off the eager execution in TF2 by running \texttt{tf.compat.v1.disable\_eager\_execution()} for this baseline.
    \item[SNS]: We run SNS for 200 steps for all datasets and project the counterfactual back to a $\ell_2$ ball. The size of the ball is set to be $0.8$ multiplied by the largest size of the ball used for the baseline Min-$\epsilon$ PGD. For Max $\ell_1$/$\ell_2$ without a norm bound, we use the norm bound from Min-$\epsilon$ PGD. Similarly, the step size is set to $2 * 0.8 * \epsilon / 200$. 
    \end{description}

    \begin{table}[t]
        \resizebox{\textwidth}{!}{%
            \begin{tabular}{c|cc|cc|cc|cc|cc|cc}
                \multicolumn{13}{c}{} \\
                \multicolumn{13}{c}{\textit{Hyper-parameters and Success Rate}} \\
                \toprule 
                \textbf{Min $\ell_1$}&\multicolumn{2}{c}{German Credit} & \multicolumn{2}{c}{Seizure} & \multicolumn{2}{c}{CTG} & \multicolumn{2}{c}{Warfarin}&\multicolumn{2}{c}{HELOC}& \multicolumn{2}{c}{Taiwanese Credit}\\
                \midrule
                $\epsilon$ & \multicolumn{2}{c}{-} & \multicolumn{2}{c}{-}& \multicolumn{2}{c}{-}&\multicolumn{2}{c}{-}&\multicolumn{2}{c}{-}& \multicolumn{2}{c}{-}\\
                \texttt{step size} & \multicolumn{2}{c}{0.05} & \multicolumn{2}{c}{0.05}& \multicolumn{2}{c}{0.05}&\multicolumn{2}{c}{0.5}&\multicolumn{2}{c}{0.01}& \multicolumn{2}{c}{0.05}\\
                \texttt{success rate} & \multicolumn{2}{c}{0.35} & \multicolumn{2}{c}{0.14}& \multicolumn{2}{c}{1.00}&\multicolumn{2}{c}{1.00}&\multicolumn{2}{c}{1.00}& \multicolumn{2}{c}{1.00}\\
                \bottomrule\\
                \textbf{Min $\ell_2$}&\multicolumn{2}{c}{German Credit} & \multicolumn{2}{c}{Seizure} & \multicolumn{2}{c}{CTG} & \multicolumn{2}{c}{Warfarin}&\multicolumn{2}{c}{HELOC}& \multicolumn{2}{c}{Taiwanese Credit}\\
                \toprule
                $\epsilon$ & \multicolumn{2}{c}{-} & \multicolumn{2}{c}{-}& \multicolumn{2}{c}{-}&\multicolumn{2}{c}{-}&\multicolumn{2}{c}{-}& \multicolumn{2}{c}{-}\\
                \texttt{step size} & \multicolumn{2}{c}{0.01} & \multicolumn{2}{c}{0.01}& \multicolumn{2}{c}{0.01}&\multicolumn{2}{c}{0.01}&\multicolumn{2}{c}{0.01}& \multicolumn{2}{c}{0.01}\\
                \texttt{success rate} & \multicolumn{2}{c}{0.84} & \multicolumn{2}{c}{0.71}& \multicolumn{2}{c}{1.00}&\multicolumn{2}{c}{1.00}&\multicolumn{2}{c}{1.00}& \multicolumn{2}{c}{1.00}\\
                \bottomrule\\
                \textbf{Min $\epsilon$ PGD}&\multicolumn{2}{c}{German Credit} & \multicolumn{2}{c}{Seizure} & \multicolumn{2}{c}{CTG} & \multicolumn{2}{c}{Warfarin}&\multicolumn{2}{c}{HELOC}& \multicolumn{2}{c}{Taiwanese Credit}\\
                \toprule
                Max. $\epsilon$ & \multicolumn{2}{c}{3.00} & \multicolumn{2}{c}{3.00}& \multicolumn{2}{c}{0.20}&\multicolumn{2}{c}{0.50}&\multicolumn{2}{c}{2.10}& \multicolumn{2}{c}{5.00}\\
                \texttt{step size} & \multicolumn{2}{c}{\texttt{adp.}} & \multicolumn{2}{c}{\texttt{adp.}}& \multicolumn{2}{c}{\texttt{adp.}}&\multicolumn{2}{c}{\texttt{adp.}}&\multicolumn{2}{c}{\texttt{adp.}}& \multicolumn{2}{c}{\texttt{adp.}}\\
                \texttt{success rate} & \multicolumn{2}{c}{0.90} & \multicolumn{2}{c}{0.86}& \multicolumn{2}{c}{0.51}&\multicolumn{2}{c}{0.85}&\multicolumn{2}{c}{1.00}& \multicolumn{2}{c}{1.00}\\
                \multicolumn{13}{c}{} \\
                \bottomrule\\
                \textbf{Looveren et al.}&\multicolumn{2}{c}{German Credit} & \multicolumn{2}{c}{Seizure} & \multicolumn{2}{c}{CTG} & \multicolumn{2}{c}{Warfarin}&\multicolumn{2}{c}{HELOC}& \multicolumn{2}{c}{Taiwanese Credit}\\
                \toprule
                $\epsilon$ & \multicolumn{2}{c}{-} & \multicolumn{2}{c}{-}& \multicolumn{2}{c}{-}&\multicolumn{2}{c}{-}&\multicolumn{2}{c}{-}& \multicolumn{2}{c}{-}\\
                \texttt{step size} & \multicolumn{2}{c}{-} & \multicolumn{2}{c}{-}& \multicolumn{2}{c}{-}&\multicolumn{2}{c}{-}&\multicolumn{2}{c}{-}& \multicolumn{2}{c}{-}\\
                \texttt{success rate} & \multicolumn{2}{c}{1.0} & \multicolumn{2}{c}{1.0}& \multicolumn{2}{c}{1.0}&\multicolumn{2}{c}{1.0}&\multicolumn{2}{c}{1.0}& \multicolumn{2}{c}{1.0}\\
                \multicolumn{13}{c}{} \\
                \bottomrule\\
                \textbf{Pawelczyk et al.}&\multicolumn{2}{c}{German Credit} & \multicolumn{2}{c}{Seizure} & \multicolumn{2}{c}{CTG} & \multicolumn{2}{c}{Warfarin}&\multicolumn{2}{c}{HELOC}& \multicolumn{2}{c}{Taiwanese Credit}\\
                \toprule
                $\epsilon$ & \multicolumn{2}{c}{0.3} & \multicolumn{2}{c}{1.0}& \multicolumn{2}{c}{1.0}&\multicolumn{2}{c}{1.0}&\multicolumn{2}{c}{1.0}& \multicolumn{2}{c}{1.0}\\
                \texttt{step size} & \multicolumn{2}{c}{-} & \multicolumn{2}{c}{-}& \multicolumn{2}{c}{-}&\multicolumn{2}{c}{-}&\multicolumn{2}{c}{-}& \multicolumn{2}{c}{-}\\
                \texttt{success rate} & \multicolumn{2}{c}{0.38} & \multicolumn{2}{c}{1.00}& \multicolumn{2}{c}{0.87}&\multicolumn{2}{c}{0.72}&\multicolumn{2}{c}{0.76}& \multicolumn{2}{c}{0.14}\\
                \multicolumn{13}{c}{} \\
            \end{tabular}%
            }
            \caption{Hyper-parameters and Success Rate for each baseline methods. \texttt{adp.} denotes that the step size for each iteration is $2*\epsilon/$\texttt{max\_steps}.}
            \label{table_param}
            \end{table}

% \subsection{Sensitivity of $a$ in SNS}\label{appendix:experiment-detail-sensitivity-of-alpha}

% We show the plot of invalidation rate (IV) of SNS+PGD on German Credit and Seizure in Fig.~\ref{fig:stab} against the change of $a \in [0, 0.9]$. We find IV is not very sensitive to the choice of $a$ so $a=0.5$ is a natural choice in our experiments.

% \begin{figure}[t]
%     \centering
%     \includegraphics[width=\textwidth]{images/Stab.pdf}
%     \caption{Stability Test of the choice of $a$ against Invalidation Rate (IV) for German Credit and Seizure.}
%     \label{fig:stab}
% \end{figure}

\subsection{Details of Retraining}\label{appendix:experiment-detail-retraining}

We evaluate counterfactual invalidation over models with one-point differences in their training set, or different random initialization. 
For each dataset, we train a base model $F(\theta)$ with a specified random seed to determine initialization, and a specified train-validation split. We use this to generate all counterfactuals.
We then train 100 models with one-point differences in the training set from a base model, as well as 100 models trained with different random initialization parameters. 
To do this, we randomly derive: a training set $S$, a set $O \subseteq S$ of size 100 that consists of points drawn randomly from test data (i.e. with which to create 100 different training sets with one point removed, $S^{(\setminus i)}$), and a test set. 
Then, For each $z_i \in O$, we train $F(\theta)'$ on $S^{(\setminus i)}$ by removing $z_i$ from $S$. 
%These are the models across which we use to estimate invalidation rate over one-point changes to the training set. We also train $100$ models with different random seeds than the baseline model, which we use to estimate invalidation rate over changes to random initialization of the model.
To train the 100 models with different initialization parameters, we simply change the numpy random seed directly before initializing a model.

\subsection{Full results of IV with Standard Deviations}\label{appendix:error-bars}

The full results of invalidation rates with standard deviations are shown in Table~\ref{table_IV_std_rs}.

\begin{table}[t]
    \resizebox{\textwidth}{!}{%
        \begin{tabular}{c|cc|cc|cc|cc|cc|cc}
            \multicolumn{13}{c}{} \\
            \multicolumn{13}{c}{\textit{Invalidation Rate (LOO)}} \\
            \toprule 
            \textbf{Method}&\multicolumn{2}{c}{German Credit} & \multicolumn{2}{c}{Seizure} & \multicolumn{2}{c}{CTG} & \multicolumn{2}{c}{Warfarin}&\multicolumn{2}{c}{HELOC}& \multicolumn{2}{c}{Taiwanese Credit}\\
            \midrule
            Min. $\ell_1$ & \multicolumn{2}{c}{0.41$\pm$0.04} & \multicolumn{2}{c}{-}& \multicolumn{2}{c}{0.07$\pm$0.09}&\multicolumn{2}{c}{0.44$\pm$0.02}&\multicolumn{2}{c}{0.30$\pm$0.03}& \multicolumn{2}{c}{0.30$\pm$0.03}\\
            $\quad \quad$+SNS & \multicolumn{2}{c}{0.00$\pm$0.00} & \multicolumn{2}{c}{-}& \multicolumn{2}{c}{0.00$\pm$0.00}&\multicolumn{2}{c}{0.00$\pm$0.00}&\multicolumn{2}{c}{0.00$\pm$0.00}& \multicolumn{2}{c}{0.00$\pm$0.00}\\
            Min. $\ell_2$ & \multicolumn{2}{c}{0.36$\pm$0.05} & \multicolumn{2}{c}{0.64$\pm$0.06}& \multicolumn{2}{c}{0.48$\pm$0.17}&\multicolumn{2}{c}{0.35$\pm$0.02}&\multicolumn{2}{c}{0.55$\pm$0.05}& \multicolumn{2}{c}{0.27$\pm$0.05}\\
            $\quad \quad$+SNS & \multicolumn{2}{c}{0.00$\pm$0.00} & \multicolumn{2}{c}{0.02$\pm$0.02}& \multicolumn{2}{c}{0.00$\pm$0.00}&\multicolumn{2}{c}{0.00$\pm$0.00}&\multicolumn{2}{c}{0.00$\pm$0.00}& \multicolumn{2}{c}{0.00$\pm$0.00}\\
            Min. $\epsilon$ PGD& \multicolumn{2}{c}{0.28$\pm$0.03} & \multicolumn{2}{c}{0.94$\pm$0.01}& \multicolumn{2}{c}{0.04$\pm$0.03}&\multicolumn{2}{c}{0.10$\pm$0.01}&\multicolumn{2}{c}{0.04$\pm$.0.01}& \multicolumn{2}{c}{0.04$\pm$0.00}\\
            % CW &  &  & && &\\
            $\quad \quad$+SNS & \multicolumn{2}{c}{0.00$\pm$0.00} & \multicolumn{2}{c}{0.04$\pm$0.02}& \multicolumn{2}{c}{0.00$\pm$0.00}&\multicolumn{2}{c}{0.01$\pm$0.00}&\multicolumn{2}{c}{0.00$\pm$0.00}& \multicolumn{2}{c}{0.00$\pm$0.00}\\
            Looveren et al. & \multicolumn{2}{c}{0.25$\pm$0.03} & \multicolumn{2}{c}{0.48$\pm$0.04}& \multicolumn{2}{c}{0.11$\pm$0.08}&\multicolumn{2}{c}{0.26$\pm$0.02}&\multicolumn{2}{c}{0.25$\pm$0.03}& \multicolumn{2}{c}{0.29$\pm$0.06}
            \\
            Pawelczyk et al. & \multicolumn{2}{c}{0.20$\pm$0.13} & \multicolumn{2}{c}{0.16$\pm$0.14}& \multicolumn{2}{c}{0.00$\pm$0.00}&\multicolumn{2}{c}{0.02$\pm$0.00}&\multicolumn{2}{c}{0.05$\pm$0.06}& \multicolumn{2}{c}{0.02$\pm$0.01}\\
            \multicolumn{13}{c}{} \\
            \multicolumn{13}{c}{\textit{Invalidation Rate (RS)}} \\
            \toprule \\
            Min. $\ell_1$ & \multicolumn{2}{c}{0.56$\pm$0.05} & \multicolumn{2}{c}{-}& \multicolumn{2}{c}{0.29$\pm$0.09}&\multicolumn{2}{c}{0.35$\pm$0.08}&\multicolumn{2}{c}{0.43$\pm$0.07}& \multicolumn{2}{c}{0.78$\pm$0.06}\\
            $\quad \quad$+SNS & \multicolumn{2}{c}{0.07$\pm$0.02} & \multicolumn{2}{c}{-}& \multicolumn{2}{c}{0.01$\pm$0.00}&\multicolumn{2}{c}{0.00$\pm$0.00}&\multicolumn{2}{c}{0.00$\pm$0.00}& \multicolumn{2}{c}{0.04$\pm$0.02}\\
            Min. $\ell_2$ & \multicolumn{2}{c}{0.56$\pm$0.06} & \multicolumn{2}{c}{0.77$\pm$0.12}& \multicolumn{2}{c}{0.49$\pm$0.15}&\multicolumn{2}{c}{0.30$\pm$0.05}&\multicolumn{2}{c}{0.61$\pm$0.07}& \multicolumn{2}{c}{0.72$\pm$0.07}\\
            $\quad \quad$+SNS & \multicolumn{2}{c}{0.06$\pm$0.04} & \multicolumn{2}{c}{0.13$\pm$0.08}& \multicolumn{2}{c}{0.00$\pm$0.00}&\multicolumn{2}{c}{0.00$\pm$0.00}&\multicolumn{2}{c}{0.00$\pm$0.00}& \multicolumn{2}{c}{0.04$\pm$0.04}\\
            Min. $\epsilon$ PGD& \multicolumn{2}{c}{0.61$\pm$0.04} & \multicolumn{2}{c}{0.94$\pm$0.12}& \multicolumn{2}{c}{0.09$\pm$0.04}&\multicolumn{2}{c}{0.12$\pm$0.03}&\multicolumn{2}{c}{0.11$\pm$0.02}& \multicolumn{2}{c}{0.24$\pm$0.07}\\
            % CW &  &  & && &\\
            $\quad \quad$+SNS & \multicolumn{2}{c}{0.12$\pm$0.03} & \multicolumn{2}{c}{0.16$\pm$0.08}& \multicolumn{2}{c}{0.00$\pm$0.00}&\multicolumn{2}{c}{0.02$\pm$0.01}&\multicolumn{2}{c}{0.00$\pm$0.00}& \multicolumn{2}{c}{0.11$\pm$0.05}\\
            Looveren et al.  & \multicolumn{2}{c}{0.40$\pm$0.03} & \multicolumn{2}{c}{0.54$\pm$0.05}& \multicolumn{2}{c}{0.18$\pm$0.08}&\multicolumn{2}{c}{0.25$\pm$0.02}&\multicolumn{2}{c}{0.34$\pm$0.05}& \multicolumn{2}{c}{0.53$\pm$0.06}\\
            Pawelczyk et al. & \multicolumn{2}{c}{0.35$\pm$0.16} & \multicolumn{2}{c}{0.11$\pm$0.17}& \multicolumn{2}{c}{0.06$\pm$0.04}&\multicolumn{2}{c}{0.01$\pm$0.00}&\multicolumn{2}{c}{0.15$\pm$0.21}& \multicolumn{2}{c}{0.20$\pm$0.09}\\
            \multicolumn{13}{c}{} \\
        \end{tabular}%
        }
        \caption{Invalidation Rates with standard deviations for each datasets and each re-training situations. Results are aggregated over 100 models.}
        \label{table_IV_std_rs}
        \end{table}

\begin{table}[t]
    \resizebox{\textwidth}{!}{%
        \begin{tabular}{c|cc|cc|cc|cc|cc|cc}
            \multicolumn{13}{c}{} \\
            \multicolumn{13}{c}{\emph{Counterfactual Cost ($\ell_2$)}} \\
            \toprule 
            \textbf{Method}&\multicolumn{2}{c}{German Credit} & \multicolumn{2}{c}{Seizure} & \multicolumn{2}{c}{CTG} & \multicolumn{2}{c}{Warfarin}&\multicolumn{2}{c}{HELOC}& \multicolumn{2}{c}{Taiwanese Credit}\\
            \midrule
            Min. $\ell_1$ & \multicolumn{2}{c}{1.33$\pm$1.07} & \multicolumn{2}{c}{-}& \multicolumn{2}{c}{0.17$\pm$0.12}&\multicolumn{2}{c}{0.50$\pm$0.33}&\multicolumn{2}{c}{0.24$\pm$0.18}& \multicolumn{2}{c}{1.56$\pm$0.94}\\
            Min. $\ell_2$ & \multicolumn{2}{c}{4.49$\pm$1.90} & \multicolumn{2}{c}{8.23$\pm$2.27}& \multicolumn{2}{c}{0.06$\pm$0.04}&\multicolumn{2}{c}{0.54$\pm$0.57}&\multicolumn{2}{c}{0.11$\pm$0.08}& \multicolumn{2}{c}{2.65$\pm$1.08}\\
            Looveren et al. & \multicolumn{2}{c}{5.37$\pm$2.53} & \multicolumn{2}{c}{8.40$\pm$6.96}& \multicolumn{2}{c}{0.11$\pm$0.06}&\multicolumn{2}{c}{1.03$\pm$0.46}&\multicolumn{2}{c}{0.45$\pm$0.45}& \multicolumn{2}{c}{2.82$\pm$1.89}\\
            Min. $\epsilon$ PGD & \multicolumn{2}{c}{1.02$\pm$0.57} & \multicolumn{2}{c}{1.36$\pm$0.38}& \multicolumn{2}{c}{0.08$\pm$0.03}&\multicolumn{2}{c}{0.31$\pm$ 0.12}&\multicolumn{2}{c}{0.32$\pm$0.12}& \multicolumn{2}{c}{0.75$\pm$0.27}\\
            \bottomrule
            Min.$\ell_1$ + SNS &\multicolumn{2}{c}{3.40$\pm$0.82} & \multicolumn{2}{c}{-}& \multicolumn{2}{c}{0.25$\pm$0.08}&\multicolumn{2}{c}{0.80$\pm$0.29}&\multicolumn{2}{c}{1.71$\pm$0.12}& \multicolumn{2}{c}{3.50$\pm$0.91}\\
            Min.$\ell_2$ + SNS &\multicolumn{2}{c}{6.23$\pm$1.65} & \multicolumn{2}{c}{9.60$\pm$2.31}& \multicolumn{2}{c}{0.21$\pm$0.04}&\multicolumn{2}{c}{0.90$\pm$0.54}&\multicolumn{2}{c}{1.71$\pm$0.11}& \multicolumn{2}{c}{4.68$\pm$1.03}\\
            PGD + SNS &\multicolumn{2}{c}{3.03$\pm$0.38} & \multicolumn{2}{c}{3.60$\pm$0.59}& \multicolumn{2}{c}{0.22$\pm$0.04}&\multicolumn{2}{c}{0.50$\pm$0.11}&\multicolumn{2}{c}{1.79$\pm$0.15}& \multicolumn{2}{c}{2.78$\pm$0.49}\\
            % CW &  &  & && &\\
            % CW+SNS & \multicolumn{2}{c}{\textbf{2.63}} & \multicolumn{2}{c}{\textbf{6.01}}& \multicolumn{2}{c}{-}&\multicolumn{2}{c}{-}&\multicolumn{2}{c}{1.75}& \multicolumn{2}{c}{3.37} \\
            Pawelczyk et al. & \multicolumn{2}{c}{7.15$\pm$2.12} & \multicolumn{2}{c}{13.66$\pm$7.46}& \multicolumn{2}{c}{1.07$\pm$0.20}&\multicolumn{2}{c}{2.62$\pm$0.79}&\multicolumn{2}{c}{1.35$\pm$0.93}& \multicolumn{2}{c}{4.24$\pm$2.23}\\
            \bottomrule
            \multicolumn{13}{c}{} \\
            \multicolumn{13}{c}{\emph{Counterfactual Cost ($\ell_1$)}} \\
            \toprule 
            \textbf{Method}&\multicolumn{2}{c}{German Credit} & \multicolumn{2}{c}{Seizure} & \multicolumn{2}{c}{CTG} & \multicolumn{2}{c}{Warfarin}&\multicolumn{2}{c}{HELOC}& \multicolumn{2}{c}{Taiwanese Credit}\\
            \midrule
            Min. $\ell_1$ & \multicolumn{2}{c}{2.09$\pm$2.00} & \multicolumn{2}{c}{-}& \multicolumn{2}{c}{0.16$\pm$0.19}&\multicolumn{2}{c}{0.48$\pm$0.59}&\multicolumn{2}{c}{0.35$\pm$0.30}& \multicolumn{2}{c}{2.40$\pm$2.84}\\
            Min. $\ell_2$ & \multicolumn{2}{c}{24.70$\pm$13.14} & \multicolumn{2}{c}{77.89$\pm$28.38}& \multicolumn{2}{c}{0.18$\pm$0.12}&\multicolumn{2}{c}{1.17$\pm$0.91}&\multicolumn{2}{c}{0.43$\pm$0.31}& \multicolumn{2}{c}{9.76$\pm$3.93}\\
            Looveren et al.& \multicolumn{2}{c}{15.93$\pm$8.93} & \multicolumn{2}{c}{76.99$\pm$69.93}& \multicolumn{2}{c}{0.16$\pm$0.12}&\multicolumn{2}{c}{1.75$\pm$0.99}&\multicolumn{2}{c}{0.97$\pm$1.05}& \multicolumn{2}{c}{6.11$\pm$5.12}\\

            Min. $\epsilon$ PGD & \multicolumn{2}{c}{6.31$\pm$3.56} & \multicolumn{2}{c}{14.55$\pm$4.15}& \multicolumn{2}{c}{0.30$\pm$0.12}&\multicolumn{2}{c}{1.07$\pm$0.42}&\multicolumn{2}{c}{1.45$\pm$0.54}& \multicolumn{2}{c}{3.19$\pm$1.09}\\
            \bottomrule
            % PGDs &  && && &\\
            Min.$\ell_1$ + SNS & \multicolumn{2}{c}{13.81$\pm$2.96} & \multicolumn{2}{c}{-}& \multicolumn{2}{c}{0.58$\pm$0.14}&\multicolumn{2}{c}{1.61$\pm$0.52}&\multicolumn{2}{c}{7.11$\pm$0.72}& \multicolumn{2}{c}{11.04$\pm$3.17}\\
            Min.$\ell_2$ + SNS & \multicolumn{2}{c}{34.38$\pm$11.54} & \multicolumn{2}{c}{96.08$\pm$27.80}& \multicolumn{2}{c}{0.57$\pm$0.10}&\multicolumn{2}{c}{2.17$\pm$0.95}&\multicolumn{2}{c}{7.18$\pm$0.73}& \multicolumn{2}{c}{16.63$\pm$3.76}\\
            PGD + SNS & \multicolumn{2}{c}{14.21$\pm$2.31} & \multicolumn{2}{c}{38.55$\pm$6.36}& \multicolumn{2}{c}{0.63$\pm$0.14}&\multicolumn{2}{c}{1.41$\pm$0.36}&\multicolumn{2}{c}{7.64$\pm$0.88}& \multicolumn{2}{c}{10.19$\pm$2.54}\\
            Pawelczyk et al. & \multicolumn{2}{c}{38.48$\pm$12.31} & \multicolumn{2}{c}{145.36$\pm$77.67}& \multicolumn{2}{c}{3.03$\pm$0.63}&\multicolumn{2}{c}{$\pm$6.48$\pm$2.66}&\multicolumn{2}{c}{4.51$\pm$3.52}& \multicolumn{2}{c}{12.22$\pm$7.49}\\
            \bottomrule
            \multicolumn{13}{c}{} \\
        \end{tabular}%
        }
        \caption{$\ell_1$ and $\ell_2$ costs of counterfactuals with standard deviations.}
        \label{table_IV:l1}
        \end{table}

\subsection{$\ell_1$ and $\ell_2$ results}\label{appendix:ell1-results}
The full results of $\ell_1$ and $\ell_2$ costs with standard deviations are shown in Table~\ref{table_IV:l1}.

\end{document}